\RequirePackage{fix-cm}
\documentclass[twocolumn]{svjour3}          
\smartqed  
\usepackage{natbib}
\usepackage{graphicx}%
\usepackage{multirow}%
\usepackage{amsmath,amssymb,amsfonts}%
\usepackage{mathrsfs}%
\usepackage[title]{appendix}%
\usepackage{textcomp}%
\usepackage{manyfoot}%
\usepackage{booktabs}%
\usepackage{algorithm}%
\usepackage{algorithmicx}%
\usepackage{algpseudocode}%
\usepackage{listings}%
\usepackage{xspace}

\usepackage{colortbl}
\usepackage{array} 

\usepackage{tabularx}
\usepackage{bm}
\usepackage{caption}
\usepackage{colortbl}
\newcommand{\name}{\textbf{\textsc{S3Diff}}\xspace}

\usepackage{pifont}

\usepackage[dvipsnames]{xcolor}
\usepackage{color, colortbl}
\definecolor{citecolor}{HTML}{0071bc}
\definecolor{tabhighlight}{HTML}{e5e5e5}
\usepackage[colorlinks,citecolor=citecolor]{hyperref}

\makeatletter
\renewcommand\paragraph{
  \@startsection{paragraph} 
  {4} 
  {\z@} 
  {.5em \@plus1ex \@minus.2ex} 
  {-.5em} 
  {\normalfont\normalsize\bfseries} 
}
\makeatother

\begin{document}

\title{Degradation-Guided One-Step Image Super-Resolution with Diffusion Priors}



\author{Aiping Zhang* \and
        Zongsheng Yue* \and
        Renjing Pei \and
        Wenqi Ren \and \\
        Xiaochun Cao
}


\institute{Aiping Zhang \at
              School of Cyber Science and Technology, Shenzhen Campus of Sun Yat-sen University, Shenzhen, China \\
              \email{zhangaip7@mail2.sysu.edu.cn}
           \and
           Zongsheng Yue \at
              S-Lab, Nanyang Technological University, Singapore \\
              \email{zsyzam@gmail.com}
           \and
           Renjing Pei \at
              Huawei Noah’s Ark Lab \\
              \email{peirenjing@huawei.com}
           \and
           Wenqi Ren \at
              School of Cyber Science and Technology, Shenzhen Campus of Sun Yat-sen University, Shenzhen, China \\
              \email{renwq3@mail.sysu.edu.cn}
           \and
           Xiaochun Cao  \at
              School of Cyber Science and Technology, Shenzhen Campus of Sun Yat-sen University, Shenzhen, China \\
              \email{caoxiaochun@mail.sysu.edu.cn}
           \and
           * Equal Contribution.
}

\date{Received: date / Accepted: date}

\maketitle

\abstract{Diffusion-based image super-resolution (SR) methods have achieved remarkable success by leveraging large pre-trained text-to-image diffusion models as priors. However, these methods still face two challenges: the requirement for dozens of sampling steps to achieve satisfactory results, which limits efficiency in real scenarios, and the neglect of degradation models, which are critical auxiliary information in solving the SR problem. In this work, we introduced a novel one-step SR model, which significantly addresses the efficiency issue of diffusion-based SR methods. Unlike existing fine-tuning strategies, we designed a degradation-guided Low-Rank Adaptation (LoRA) module specifically for SR, which corrects the model parameters based on the pre-estimated degradation information from low-resolution images. This module not only facilitates a powerful data-dependent or degradation-dependent SR model but also preserves the generative prior of the pre-trained diffusion model as much as possible. Furthermore, we tailor a novel training pipeline by introducing an online negative sample generation strategy. Combined with the classifier-free guidance strategy during inference, it largely improves the perceptual quality of the super-resolution results. Extensive experiments have demonstrated the superior efficiency and effectiveness of the proposed model compared to recent state-of-the-art methods. Code is available at \url{https://github.com/ArcticHare105/S3Diff}}




\maketitle

\section{Introduction} \label{sec:intro}
Image super-resolution (SR) is a long-standing and challenging problem in computer vision, aiming to restore a high-resolution (HR) image from its low-resolution (LR) counterpart. The LR images usually suffer from various complex degradations, such as blurring, downsampling, noise corruption, etc. Even worse, the degradation process is often unknown in real-world scenarios. This inherent ambiguity in the degradation model further heightens the complexity of the SR problem, driving substantial research efforts over the past years.

\begin{figure}[t]
    \centering
    \includegraphics[width=1.0\linewidth]{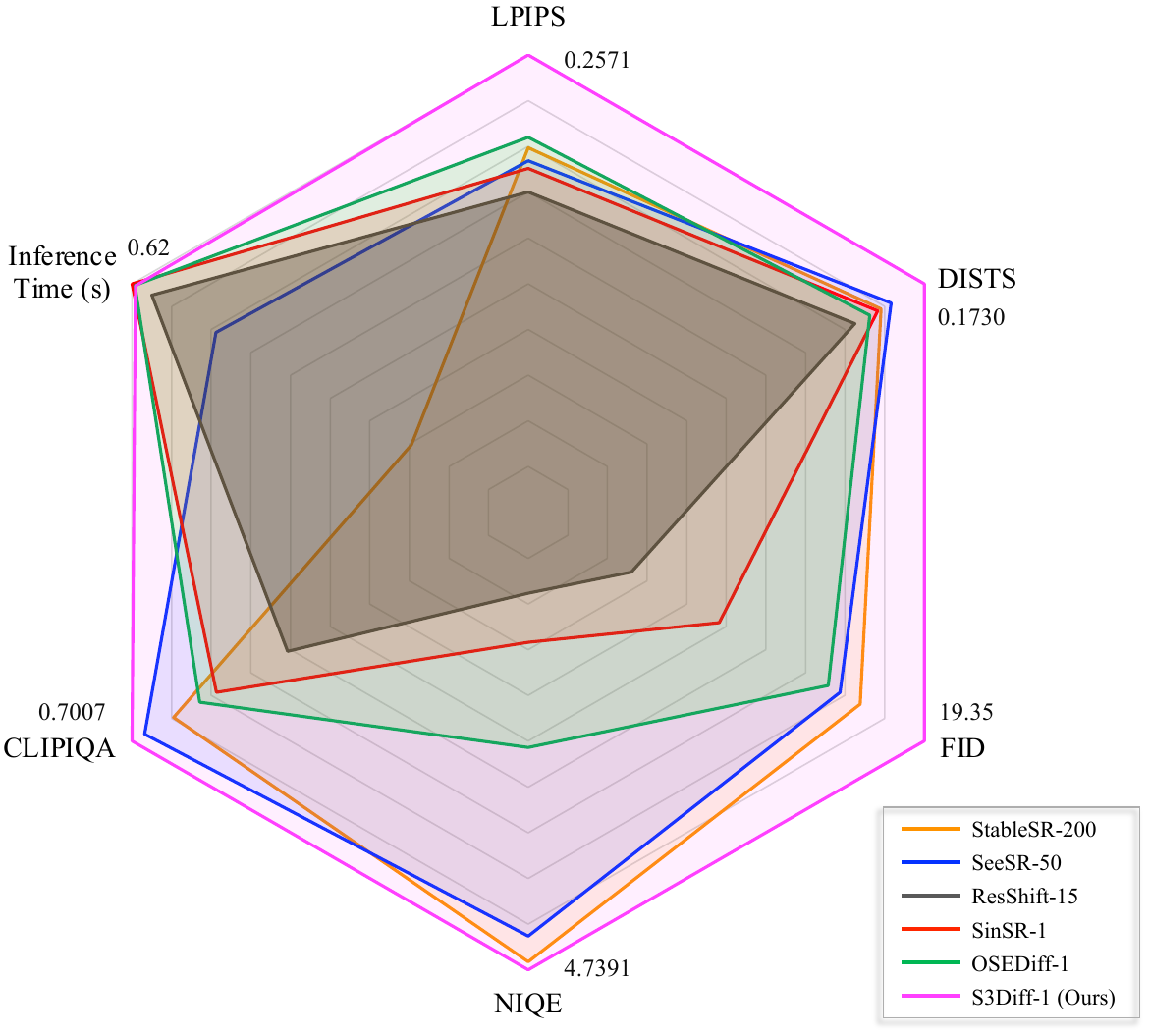}
    \caption{Comparison of performance and complexity among DM-based SR methods on the DIV2K-Val dataset~\citep{div2k}. Metrics like LPIPS, DISTS, NIQE, FID, and inference time, where smaller scores indicate better image quality, are inverted. All metrics are normalized for better visualization. \name{} attains top-tier performance in both image quality and complexity with just a single forward pass.}
    \vspace{-3mm}
    \label{fig:radar_figure}
\end{figure}

Diffusion models have emerged as a formidable class of generative models, particularly excelling in image generation tasks. Building on the foundational work \citep{sohl2015deep}, these models have significantly advanced, resulting in highly effective frameworks \citep{ho2020denoising,song2020score}. The field of SR has particularly benefited from the diffusion models due to their ability to capture fine-grained details and generate high-fidelity images. Current diffusion-based SR approaches can be broadly classified into two categories. The first category involves the specific redesign of diffusion models for SR, including SR3 \citep{saharia2022image}, SRDiff \citep{li2022srdiff}, ResShift \citep{yue2023resshift,yue2024efficient}, and others  \citep{luo2023image,delbracio2023inversion,xia2023diffir,wang2023sinsr}. Motivated by the huge success of large text-to-image (T2I) models, the second category harnesses a large pre-trained T2I model, like Stable Diffusion (SD) \citep{rombach2022high}, as a prior to facilitate the SR task. Following the pioneering work of StableSR \citep{wang2023exploiting}, several relevant studies \citep{lin2023diffbir,yang2023pixel,yu2024scaling,wu2023seesr} have recently emerged. These methods, which are built upon T2I models trained with hundreds to thousands of diffusion steps, typically require dozens of sampling steps even after acceleration, limiting their inference efficiency. While some methods \citep{yue2023resshift,wang2023sinsr} in the first category can significantly reduce sampling steps by designing a shorter diffusion trajectory, they necessitate training the model from scratch and cannot capitalize on the extensive knowledge embedded in large pre-trained T2I models. 

Recently, the acceleration of diffusion models has attracted much attention. Using distillation strategies~\citep{salimans2022progressive,sauer2023adversarial,yin2024one,sauer2024fast}, efficient samplers~\citep{ddim,lu2022dpm,lu2022dpmplus} and straight forward path~\citep{lipmanflow,liu2022flow} effectively reduce inference steps and achieve promising generation quality. However, directly applying these approaches for efficient SR could be problematic. Unlike text-to-image generation, super-resolution relies on an LR input image to create an HR image. The LR image provides more detailed content for the target image than textual descriptions. Moreover, understanding the process of degradation is essential in generating high-resolution images, and when used effectively, it can positively influence the SR process. 

Considering these observations, this work follows the second research line, focusing on efficiently leveraging LR inputs and degradation guidance to better harness the T2I prior for effective and efficient super-resolution. Specifically, we propose a Single-Step Super-resolution Diffusion network (\name) for addressing the problem of real-world SR. Benefiting from advances in accelerating diffusion models, we take advantage of the T2I prior of SD-Turbo \citep{sauer2023adversarial} due to its efficient few-step inference and powerful generative capabilities. Inspired by recent methods like DifFace~\citep{yue2022difface} and Diff-SR~\citep{li2023dissecting}, which suggest that LR images provide a robust and effective starting point for the diffusion reverse process, we use the LR image with slight or no noise perturbation as input to maximize the retention of semantic content. To achieve high-quality super-resolution, we integrate the T2I prior with Low-Rank Adaptation (LoRA)~\citep{hu2022lora}, transforming it into a one-step SR model that maintains its generative capabilities. In comparison to previous fine-tuning approaches ~\citep{wang2023exploiting,lin2023diffbir,yu2024scaling}, the application of LoRA offers a more lightweight and rapidly adaptable solution for the SR task.
\begin{figure}[!t]
    \centering
    \includegraphics[width=\linewidth]{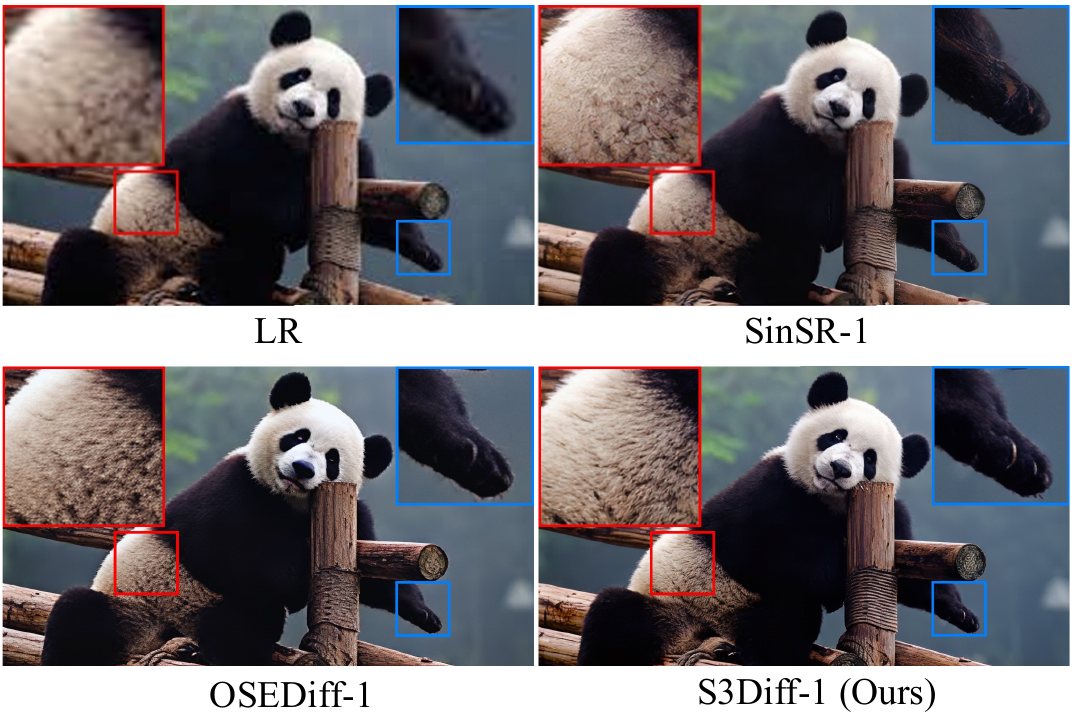}
    \caption{Qualitative comparisons on one typical real-world example of the proposed method and the most recent state-of-the-arts, including SinSR \citep{wang2023sinsr} and OSEDiff~\citep{wu2024one}. (\textbf{Zoom in for details})}\label{fig:comparison_panda}
\end{figure}

In addition to the straightforward fine-tuning strategy based on the naive LoRA, we advance our approach specifically for SR. Considering the pivotal role of the degradation model in addressing the SR problem~\citep{zhang2018learning,mou2022metric,yue2024deep}, we design a degradation-guided LoRA module that effectively leverages degraded information from LR images. This module draws on the core principles of LoRA, which involves a modification of the targeted parameter $W\in\mathcal{R}^{d\times n}$ via a residual decomposition, namely $\bm{W}_{\text{new}}=\bm{W}+\bm{A}\bm{B}$, where $\bm{A}=[\bm{a}_1,\cdots,\bm{a}_r]\in\mathcal{R}^{d\times r}$ and $\bm{B}\in\mathcal{R}^{r\times n}$ are low-rank matrices. From the perspective of mathematics, the update for $\bm{W}$ is implemented in a low-dimensional space spanned by $\{\bm{a}_1,\cdots,\bm{a}_r\}$, determined by $\bm{A}$, with $\bm{B}$ controlling the update directions. To better adapt LoRA to SR, we pre-estimate the degraded information using the degradation estimation model from~\citep{mou2022metric} and then use this information to modulate $B$, thereby refining the update directions of $\bm{W}$. This degradation-guided LoRA module is appealing in two aspects. On the one hand, it facilitates a data-dependent model wherein parameters are adaptively modified based on the specific degraded information from the LR image. On the other hand, during the testing phase, the degraded information can either be predicted by the degradation estimated model or manually set by users, enabling an interactive interface between the SR model and the user. To further enhance perceptual quality, we develop a novel training pipeline by introducing an online negative sample generation strategy. This approach makes full use of the LR image to align poor-quality concepts with negative prompts, enabling classifier-free guidance~\citep{ho2022classifier} during inference to further improve visual effects.
As shown in Figure \ref{fig:radar_figure} and Figure \ref{fig:comparison_panda}, \name{} can produce high-quality HR images with enhanced fidelity and perceptual quality in a single forward pass, while significantly reducing inference time and requiring only a few trainable parameters.

In summary, our contributions are as follows:
\begin{itemize}
    \item We propose a Single-Step Super-resolution Diffusion model (\name), which leverages the T2I prior from SD-Turbo~\citep{sauer2023adversarial}, achieving high-quality super-resolution with significantly reduced inference time and minimal trainable parameters.
    \item We introduce a novel degradation-guided LoRA module that adaptively modifies model parameters based on specific degraded information, extracted from the LR images or provided by the user, enhancing the SR process with a user-interactive interface.
    \item We develop an innovative training pipeline with online negative sample generation, aligning low-quality concepts with negative prompts to enable classifier-free guidance, significantly improving visual effects in generated HR images.
\end{itemize}




\section{Related Work} \label{sec:related_work}

Image super-resolution has garnered increasing attention, evolving from traditional Maximum A Posteriori (MAP)-based methods to advanced deep learning techniques. MAP-based approaches emphasize the manual design of image priors to guide restoration, focusing on non-local similarity~\citep{dong2012nonlocally,zhang2012single}, low-rankness~\citep{dong2013nonlocal,gu2015convolutional}, and sparsity~\citep{yang2010image,kim2010single}, among others~\citep{sun2008image,huang2015single}. These methods rely on mathematical models to impose constraints that help reconstruct high-resolution images from low-resolution inputs. Deep learning-based methods, on the other hand, leverage large datasets to train neural networks that can directly map low-resolution images to high-resolution counterparts. Since the introduction of SRCNN~\citep{dong2014learning}, various approaches have emerged, focusing on aspects such as network architectures~\citep{shi2016real,zhang2017beyond,zhang2018learning,liang2021swinir,chen2023activating}, which improve feature extraction and representation, and loss functions~\citep{johnson2016perceptual,lpips,wang2018esrgan,yue2024deep}, which enhance visual quality by prioritizing perceptual fidelity. Additionally, degradation models~\citep{wang2021real,zhang2021designing} and image prior integration~\citep{pan2021exploiting,chen2022femasr} further refine the SR process by combining traditional strengths with deep learning. Recently, diffusion-based SR techniques have gained prominence, categorized into model-driven and prior-driven methods: model-driven approaches leverage specific architectures and training processes incorporating diffusion mechanisms, while prior-driven methods utilize statistical properties of natural images to guide diffusion, producing realistic and detailed high-resolution outputs. Given that our work falls within the diffusion-based methods, we provide a brief overview of these approaches.

\subsection{Model-driven methods}

Model-driven approaches focus on tailoring a diffusion model specifically for super-resolution. A direct method involves modifying the inverse sampling process to include low-resolution images as conditions, followed by retraining the model from scratch. This is exemplified in works, such as SR3 \citep{saharia2022image} and SRDiff \citep{li2022srdiff}.
SR3 \citep{saharia2022image} introduces a conditional diffusion model that incorporates LR images to guide the generation of HR outputs. It emphasizes training with a large dataset to enhance the model's ability to produce finer details in the HR images.
Building on this, SRDiff \citep{saharia2022image} mainly focuses on refining the noise scheduling process, allowing for more precise control over the denoising steps.
IDM \citep{gao2023implicit} introduces an implicit neural representation into the diffusion model framework to tackle continuous SR tasks. This method allows the model to adaptively learn representations that capture the nuances of various resolutions, improving its flexibility and performance on diverse datasets.
Additionally, ResShift \citep{yue2023resshift,yue2024efficient}, which builds up a shorter Markov chain between the LR image and its corresponding HR image within the discrete framework of DDPM \citep{ho2020denoising}, significantly reducing the sampling steps during inference.
Concurrently, IR-SDE \citep{luo2023image} and InDI \citep{delbracio2023inversion} apply similar concepts within the framework of Stochastic Differential Equations (SDEs). These methods harness continuous noise processes to enhance the stability and efficiency of the diffusion models, making them robust against variations in input resolution.
Furthermore, SinSR \citep{wang2023sinsr} proposes a one-step diffusion model by distilling the ResShift method. This innovation simplifies the diffusion process to a single step, greatly improving computational efficiency while retaining high fidelity in the generated HR images. Despite their innovations, these methods are generally trained on small-scale SR datasets. This limitation means they cannot fully exploit the rich prior knowledge found in large pre-trained T2I models.

\begin{figure*}[!t]
  \centering
  \includegraphics[width=2\columnwidth]{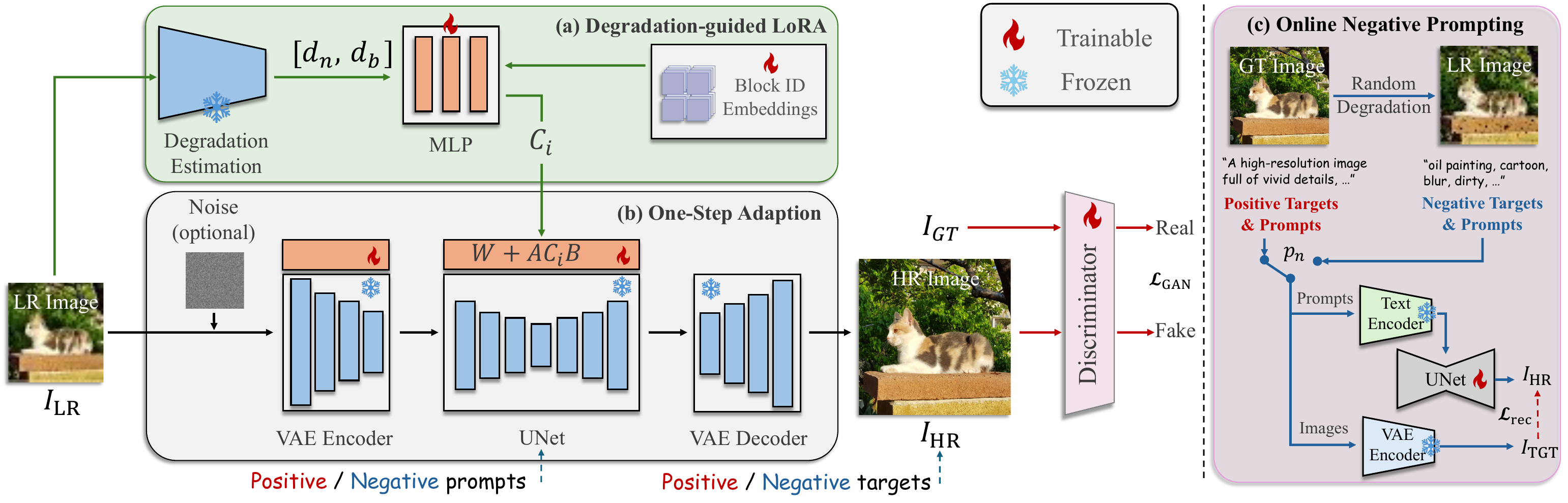}
  \caption{Overview of \name. We enhance a pre-trained diffusion model for one-step SR by injecting LoRA layers into the VAE encoder and UNet. Additionally, we employ a pre-trained Degradation Estimation Network to assess image degradation that is used to guide the LoRAs with the introduced block ID embeddings. We tailor a new training pipeline that includes an online negative prompting, reusing generated LR images with negative text prompts. The network is trained with a combination of a reconstruction loss and a GAN loss.}\label{fig:sr_framework}
\end{figure*}

\begin{figure}[!t]
  \centering
  \includegraphics[width=1\columnwidth]{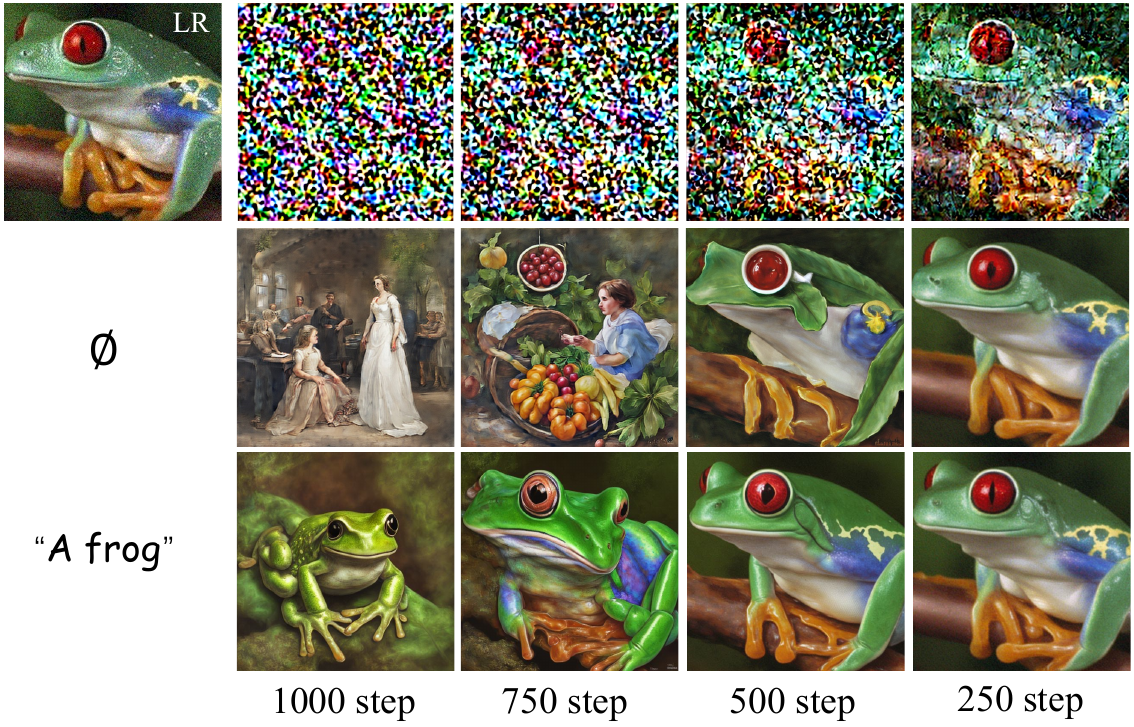}
  \caption{We demonstrate images generated from various steps using the pre-trained SD-Turbo, both with and without text prompts.}\label{fig:start_steps_comparison}
\end{figure}

\subsection{Prior-driven methods}

Motivated by the powerful image generative capabilities of large T2I models, researchers have explored the potential of leveraging these pre-trained models as priors to facilitate the SR task. Rombach \textit{et al.}~\citep{rombach2022high} propose Latent Diffusion Models (LDM), which are further used to create an upscaler for super-resolution. By operating in a compressed latent space, this method reduces computational overhead while maintaining high-quality results.
Wang \textit{et al.}~\citep{wang2023exploiting} pioneered this approach with StableSR, which generates HR images by modulating the features of the T2I model through observed LR images. It adjusts the internal representations of the model to enhance details and sharpness. Different from StableSR, DiffBIR~\citep{lin2023diffbir} offers an innovative strategy by incorporating conditional information from LR images through residual addition, inspired by DifFace~\citep{yue2022difface} and ControlNet~\citep{zhang2023adding}, enhancing the model's ability to handle complex textures and fine details.
Moreover, SUPIR~\citep{yu2024scaling} fine-tunes a large SR model derived from SDXL on an extensive dataset. It improves performance by adapting the model to diverse high-resolution data, allowing it to generalize better across various image types and text conditions. Other explorations in this domain include PASD~\citep{yang2023pixel}, CoSeR~\citep{sun2023coser}, and SeeSR~\citep{wu2023seesr}, also make some significant exploration along this research line. Unlike this fine-tuning strategy, some works suggest correcting the generated intermediate results of a pre-trained diffusion model using degradation models, such as CCDF \citep{chung2022come}, DDRM \citep{kawar2022denoising}, DDNM ~\citep{wang2022zero}, DPS \citep{chung2022diffusion}, and so on \citep{chung2022improving,song2022pseudoinverse}.
Recently, OSEDiff~\citep{wu2024one} proposes to adapt pre-trained T2I models into a one-step SR model by employing distribution matching distillation \citep{yin2024one} to maintain the fidelity of generated HR images. Despite the promising results, these methods rely on well-defined degradation models, limiting their applicability in blind SR tasks under real-world conditions.    

\section{Method} \label{sec:method}

Given a large-scale pre-trained T2I diffusion model capable of generating realistic images, our goal is to develop an efficient yet powerful SR model based on the T2I model. To achieve this, we need to address two key questions: i) under the iterative sampling framework of diffusion models, is it possible to derive a one-step SR model that extremely meets the efficiency requirement? ii) how can we effectively harness the generative prior encapsulated in the given T2I model to facilitate the SR task while minimizing the training cost? In this paper, we propose a novel Single-Step Super-resolution Diffusion network (\name), which is presented in Figure~\ref{fig:sr_framework}, to answer these questions in detail.

In this section, we first provide our solution to adapt a pre-trained T2I model to one-step SR (Section~\ref{sec:one_step_solution}),  wherein parameters are adaptively modified with our degradation-guided LoRA (Section~\ref{sec:degradation_guided_LoRA}). We also introduce a new training strategy called online negative prompting (Section~\ref{sec:online_negative_prompting}), which helps the model avoid generating low-quality images. The adopted losses are finally introduced in Section~\ref{sec:loss}.

\subsection{One-step Solution} \label{sec:one_step_solution}

Our objective is to facilitate a one-step super-resolution method using pre-trained diffusion models. This allows us to shift our focus toward leveraging high-level generative knowledge and diffusion priors embedded within the pre-trained models, rather than on the iterative denoising process. 
By doing so, we can enhance the efficiency and efficacy of the pre-trained diffusion model to produce high-quality images from LR inputs. 

We start with the selection of the pre-trained T2I base model from several prominent candidates, including PixArt~\citep{chen2023pixart}, Imagen~\citep{imagen}, IF~\citep{dif}, and SD models~\citep{rombach2022high}. The performance of the current PixArt model does not match that of the SD models, particularly the SDXL variant~\citep{podell2023sdxl}, likely due to its relatively limited number of parameters. Both Imagen and IF adopt a hierarchical generative framework, which poses challenges for adaptation to SR. Conversely, the direct generation mechanism employed by SD models is more friendly to SR. We thus focus on SD models.

In this paper, we consider SD-Turbo~\citep{sauer2023adversarial}, a distilled variant of the SD model designed to enhance sampling efficiency. SD-Turbo performs a distillation on four specific steps from the original 1000-step diffusion process.
Thus, SD-Turbo indeed acts as a robust denoiser at distinct noise levels corresponding to these four steps.
Figure \ref{fig:start_steps_comparison} demonstrates this by showing the \textbf{\textit{one-step prediction}} results of SD-Turbo using appropriate text prompts (e.g., ``\emph{A frog}'' in the figure) on the four distilled steps. This finding encourages us to adapt SD-Turbo into a single-step SR model.
However, in practical scenarios, we typically do not have access to the image description. Some methods \citep{wu2023seesr, yu2024scaling, yang2023pixel, wu2024one} use pre-trained models to generate image tags or descriptions but face issues with inaccuracy. Notably, the LR image provides more detailed content for the target image than textual descriptions. Figure \ref{fig:start_steps_comparison} shows that even without textual guidance, inputs with reduced noise produce outputs with consistent content. Furthermore, recent methods like DifFace~\citep{yue2022difface} and Diff-SR~\citep{li2023dissecting} demonstrate that LR images are effective starting points for the diffusion reverse process. Building on these findings, we directly use the LR image with little or no noise as input to maximize the retention of semantic content.

\textbf{Remark}. Notably, SinSR \citep{wang2023sinsr} is also a diffusion-based SR model enabling one-step prediction. However, it is distilled from ResShift \citep{yue2023resshift,yue2024efficient}, a relatively small SR model trained from scratch, and therefore cannot harness the rich prior knowledge embedded in large pre-trained T2I models. This work takes a step forward, aiming to develop a one-step SR model based on the powerful generative prior of large T2I models. In addition, different from OSEDiff \cite{wu2024one}, we aim to fully harness the rich content information in LR images, instead of relying on off-the-shelf methods to extract text prompts.

\subsection{Adaption Solution}

In this section, we concentrate on the fine-tuning strategies for the proposed one-step SR model, with the goal of enhancing its awareness of degradation and the LR input.
In general, we choose to fine-tune the VAE encoder and diffusion UNet. Fine-tuning the VAE encoder serves as a pre-cleaning function, aiming to better align with the original training process of T2I that was implemented on clean images. For the VAE decoder, recent approaches \citep{wang2023exploiting,parmar2024one} attempt to fine-tune it alongside the additional skip connections to maintain content consistency. However, our empirical findings suggest that freezing the VAE decoder ensures higher perceptual quality without compromising consistency, as demonstrated in Sec.~\ref{sec:ablation_study}.

Besides, unlike previous works \citep{wang2023exploiting,sun2023coser,lin2023diffbir,yu2024scaling} based on SFT~\citep{ost} or ControlNet~\citep{zhang2023adding}, we opt to inject LoRA~\citep{hu2022lora} layers into the pre-trained T2I model for efficient fine-tuning. This LoRA-based strategy not only preserves the prior knowledge embedded in the T2I model as much as possible but also significantly reduces the learnable parameters owning to the low-rank assumption. Considering the challenge of SR, which mainly arises from the complexity and unknown nature of the degradation model, incorporating auxiliary degraded information has proven beneficial \citep{zhang2018learning,mou2022metric,yue2024deep}. Building on this insight, we design a degradation-guided LoRA module to incorporate the estimated degraded information, as detailed in the following presentation.

\subsubsection{Degradation-guided LoRA} \label{sec:degradation_guided_LoRA}

Without loss of generality, we consider the fine-tuning for a specific network parameter $\bm{W}\in\mathcal{R}^{d\times n}$ in the base model. LoRA introduces a residual low-rank decomposition, namely $\bm{W}_{\text{new}}=\bm{W}+\bm{A}\bm{B}$, where $\bm{A}\in\mathcal{R}^{d\times r}$ and $\bm{B}\in\mathcal{R}^{r\times n}$ are low-rank matrices, where $r \ll d$ and $r \ll n$. By viewing $\bm{A}$ as $[\bm{a}_1,\cdots,\bm{a}_r]$ and $\bm{B}$ as $[\bm{b}_1^{T},\cdots,\bm{b}_r^{T}]^{T}$, we can rewrite the decomposition as:
\begin{equation}
    \bm{W}_{\text{new}}=\bm{W}+\bm{A}\bm{B} = \bm{W}+ \sum_{i=1}^r \bm{a}_i \bm{b}_i^T.
\end{equation}
This formulation indicates that LoRA updates $\bm{W}$
in a low-dimensional subspace spanned by $\{\bm{a}_1,\cdots,\bm{a}_r\}$, with $\bm{B}$ controlling the update directions. 
To enhance LoRA's adaptability for SR, we propose making it aware of degradation. This means LoRA is tailored specifically for each degraded image. Specifically, we first incorporate a degradation estimation model proposed by Mou~\textit{et al.}~\citep{mou2022metric}. This model unifies the image degradation into a 2-dimensional vector $\mathbf{d}=\{d_n, d_b\} \in [0,1]^2$, which quantifies the extent of \emph{noise} and \emph{blur}.
The estimated degradation vector $\mathbf{d}$ is transformed by a Gaussian Fourier embedding layer~\citep{tancik2020fourier} to enhance the model's ability to learn complex functions over continuous inputs, which can be formulated as:
\begin{equation}
    \mathbf{d}_e = \operatorname{concat}[\sin(2\pi \mathbf{d} \mathbf{W}_e^T), \cos(2\pi \mathbf{d} \mathbf{W}_e^T)],
\end{equation}
where $\mathbf{W}_e \in \mathbb{R}^{m}$ is a random-initialized matrix and $\mathbf{d}_e \in \mathbb{R}^{2 \times 2m}$, $\mathbf{d}_e$ is then fed into an MLP to generate a correction matrix $\mathbf{C}=[\bm{c}_1,\bm{c}_2,\cdots, \bm{c}_r] \in \mathbb{R}^{r \times r}$, which refines the update direction $\mathbf{B}$ as follows:
\begin{equation}
   \bm{W}_{\text{new}}=\bm{W}+\bm{A}\left(\bm{C}\bm{B}\right) = \bm{W} + \sum_{i=1}^r \bm{a}_i (\bm{c}_i^T \bm{B}).
\end{equation}
This refinement ensures that the update direction of $\bm{W}$ is degradation-aware. Such a degradation-guided LoRA module offers dual benefits. Firstly, it allows for dynamic, data-dependent adjustment of model parameters in response to the specific degradation from the LR image. Secondly, it provides a flexible way to handle the degraded information during testing, which can be automatically predicted by the degradation estimation model or manually configured by the user. 

However, using a shared $\bm{C}$ for all LoRA layers limits the flexibility of the fine-tuned model. Conversely, using a separate $\bm{C}$ for each LoRA layer necessitates a distinct MLP for each layer, significantly increasing the overall number of learnable parameters. To facilitate a flexible parameterization of degradation-guided LoRA within a pre-trained diffusion model consisting of \(L\) blocks, we introduce a set of block ID embeddings \(\mathcal{I} = \{\bm{l}_i\}_{i=1}^L\). The matrix $\bm{C}_{i}$ for the $i$-th block can be generated as:
\begin{equation}
   \bm{C}_{i} = \operatorname{MLP}(\operatorname{FC}(\mathbf{d}),\bm{l}_i).
\end{equation}
Here, we initially project the degradation estimation \(\mathbf{d}_e\) into a higher dimension to prevent it from being overwhelmed by block ID embeddings. We then feed the concatenation of $\{\operatorname{FC}(\mathbf{d}_e),\bm{l}_i\}$ to a shared MLP. This approach generates unique degradation-guided LoRA for each block in the pre-trained diffusion model. Moreover, the embeddings for block ID are learned through back-propagation, facilitating end-to-end training of the entire SR smodel.

\subsubsection{Online Negative Prompting} \label{sec:online_negative_prompting}

Adapting the T2I model to SR presents another challenge beyond input differences: the absence of a text prompt. As shown in Fig. \ref{fig:start_steps_comparison}, the pre-trained diffusion model's ability to generate high-quality outputs depends significantly on appropriate text prompts. However, we only use LR images as input, leading to a gap when adapting the T2I model to SR. Some methods~\citep{wu2023seesr,wu2024one,yu2024scaling} try to address this issue by introducing a degradation-robust prompt extractor to extract tags or employing Multi-modal Large Language Model (MLLM) to obtain dense captions from degraded images. However, text descriptions extracted from degraded images can be inaccurate, often leading diffusion models to produce inconsistent restoration results. Moreover, relying on MLLM introduces significant extra overhead over the SR model.
Actually, the input LR image already provides rich information about the image's semantic content. Therefore, beyond guiding the SR model on the image's elements, we should also focus on prompting it regarding what defines good and poor perceptual quality.
Notably, the recent method, SUPIR~\citep{yu2024scaling}, incorporates negative samples during training. However, it depends on SDXL~\citep{podell2023sdxl} to generate low-quality images offline, leading to a gap regarding the concept of ``poor quality'' between the real-world degraded images and those generated artificially, and further introducing additional training overhead. 
To help the SR model more efficiently understand the concept of ``poor quality'', we propose an online negative prompting strategy.
Specifically, during training, in each mini-batch, we randomly replace the target HR image with its synthesized LR image, using a sampling probability $p_n$, to constitute the negative target. Therefore, the target images $\mathbf{I}_{TGT}$ for supervising the model are constructed by combining the mixed LR and HR images, as shown in Figure~\ref{fig:sr_framework}.
Negative targets are associated with negative prompts like \emph{``oil painting, cartoon, blur, dirty, messy, low quality, deformation, low resolution, oversmooth''}, as used in~\citep{yu2024scaling}. Meanwhile, we use a general positive prompt \emph{``a high-resolution image full of vivid details, showcasing a rich blend of colors and clear textures''} for positive targets. In the training phase, we forward the positive/negative text prompts into the UNet, whose outputs are supervised by the corresponding positive/negative targets, facilitating the SR model being awareness of image quality.
During inference, we use Classifier-Free Guidance (CFG)~\citep{ho2022classifier} to ensure the model avoids producing low-quality images. Specifically, the SR model makes two predictions using positive prompts $t_{\mathrm{pos}}$ and negative prompts $t_{\mathrm{neg}}$, then fuses these results for the final output as follows:
\begin{equation}
\begin{aligned}
z_{\mathrm{pos}} &= \boldsymbol{\epsilon}_\theta(\mathbf{E}_\theta(I_{LR}), t_{\mathrm{pos}}), \\
z_{\mathrm{neg}} &= \boldsymbol{\epsilon}_\theta(\mathbf{E}_\theta(I_{LR}), t_{\mathrm{neg}}), \\
z_{\mathrm{out}} &= z_{\mathrm{neg}} + \lambda_{\mathrm{cfg}} (z_{\mathrm{pos}} - z_{\mathrm{neg}}),
\end{aligned}
\end{equation}
where $\mathbf{E}_\theta$ and $\boldsymbol{\epsilon}_\theta$ stand for the VAE encoder and the UNet denoiser, $\lambda_{\mathrm{cfg}}$ is the guidance scale. Different from SUPIR~\citep{yu2024scaling}, which generates negative samples offline using SDXL, we reuse synthesized LR images during training, adding no extra overhead to the training pipeline.


\subsection{Loss Functions} \label{sec:loss}
To train the model, we adopt a reconstruction loss $\mathcal{L}_\text{Rec}$, including a L2 loss $\mathcal{L}_2$ and a LPIPS loss $\mathcal{L}_\text{LPIPS}$.
Inspired by ADD \citep{sauer2023adversarial}, which leverages an adversarial distillation strategy, we also incorporate a GAN loss to minimize the distribution gap between generated images and real HR images. Since our target is to achieve a one-step SR model instead of that with four steps in \citep{sauer2023adversarial}, we can simplify ADD by removing the teacher model which supervises the intermediate diffusion results. Thus, the full learning objective can be expressed as follows:
\begin{equation}
\min_{\mathbf{G}_\theta} \lambda_\text{L2}\mathcal{L}_{2} + \lambda_\text{LPIPS}\mathcal{L}_{\text{LPIPS}} + \lambda_\text{GAN} \mathcal{L}_\text{GAN},
\end{equation} 
where $G(\cdot)$ represents the generator, namely our model, $\lambda_\text{L2}$, $\lambda_\text{LPIPS}$ and $\lambda_\text{GAN}$ are balancing weights.
The GAN loss is defined as:
\begin{equation}
    \begin{aligned}
        \mathcal{L}_\text{GAN} & =  \mathbb{E}_{I_{gt} \sim P_{\text{GT}}} \left[ \log \mathbf{D}_{\phi}(I_{gt}) \right] \\
        & + \mathbb{E}_{I_{lq} \sim P_{\text{LR}}} \left[ \log (1 - \mathbf{D}_{\phi}(\mathbf{G}_\theta(I_{lq})))\right],
    \end{aligned}
\end{equation}
where $\mathbf{D}_{\phi}$ denotes the discriminator with parameters $\phi$. We follow~\citep{kumari2022ensembling} by using a pre-trained DINO~\citep{caron2021emerging} model as a fixed backbone for the discriminator and introduce multiple independent classifiers, each corresponding to a distinct level feature of the backbone model. Notably, when using the proposed online negative prompting, we apply the GAN loss exclusively to the generated HR images that correspond to positive targets.

\begin{table*}[!t]
\setlength{\tabcolsep}{2.5pt}
\centering
\caption{Quantitative comparison with state-of-the-art methods on both synthetic and real-world benchmarks. The best and second best results are highlighted in \textcolor{red}{\textbf{red}} and \textcolor{blue}{\underline{blue}}, respectively. We report the results using publicly available codes and checkpoints of the compared methods.} \label{tab:compare_methods1}
\vspace{-2mm}
\fontsize{9pt}{12pt}\selectfont
\begin{tabular}{l|l|ccccccccc}
\toprule
\toprule
Datasets & Methods & PSNR$\uparrow$ & SSIM$\uparrow$ & LPIPS$\downarrow$ & DISTS$\downarrow$ & FID$\downarrow$ & NIQE$\downarrow$ & MANIQA$\uparrow$ & MUSIQ$\uparrow$ & CLIPIQA$\uparrow$ \\ 
\midrule
\multirow{11}{*}{\textit{DIV2K-Val}} 
& BSRGAN & \textcolor{blue}{\underline{24.58}} & \textcolor{blue}{\underline{0.6269}} & 0.3351 & 0.2275 & 44.23 & 4.7527 & 0.3560 & 61.19 & 0.5243 \\
& Real-ESRGAN& 24.29 & \textcolor{red}{\textbf{0.6371}} & 0.3112 & 0.2141 & 37.65 & \textcolor{red}{\textbf{4.6797}} & 0.3822 & 61.06 & 0.5278 \\ 
\cmidrule(lr){2-11}
& LDM-100 & 23.49 & 0.5762 & 0.3119 & 0.2727 & 41.37 & 5.0249 & \textcolor{red}{\textbf{0.5127}} & 62.27 &  0.6245 \\
& StableSR-200 & 23.28 & 0.5732 & 0.3111 & 0.2043 & \textcolor{blue}{\underline{24.31}} & 4.7570 & 0.4200 & 65.81 & 0.6753 \\
& PASD-20 & 24.32 & 0.6218 & 0.3763 & 0.2184 & 30.17 & 5.2946 & 0.4022 & 61.19 & 0.5676 \\
& DiffBIR-50  & 23.64 & 0.5647 & 0.3524 & 0.2128 & 30.72 &  \textcolor{blue}{\underline{4.7042}} & 0.4768 & 65.81 & 0.6704 \\
& SeeSR-50  & 23.67 & 0.6042 & 0.3194 &  \textcolor{blue}{\underline{0.1968}} &  25.89 & 4.8158 &  \textcolor{blue}{\underline{0.5041}} & \textcolor{red}{\textbf{68.67}} &  \textcolor{blue}{\underline{0.6932}} \\
\cmidrule(lr){2-11}
& ResShift-15 & \textcolor{red}{\textbf{24.71}} & 0.6234 & 0.3402 & 0.2245 & 42.01 & 6.4732 & 0.3985 & 60.87 & 0.5933 \\
& SinSR-1 & 24.41 & 0.6017 & 0.3244 & 0.2068 & 35.22 & 5.9996 & 0.4239 & 62.73 & 0.6468 \\
& OSEDiff-1 & 23.30 & 0.5970 & \textcolor{blue}{\underline{0.3046}} & 0.2129 & 26.80 & 5.4050 & 0.4458 & 65.56 & 0.6584 \\
& \multicolumn{1}{>{\columncolor[gray]{0.8}}l|}{S3Diff-1 (Ours)} &
\multicolumn{1}{>{\columncolor[gray]{0.8}}c}{23.40} & 
\multicolumn{1}{>{\columncolor[gray]{0.8}}c}{0.5953} & 
\multicolumn{1}{>{\columncolor[gray]{0.8}}c}{\textcolor{red}{\textbf{0.2571}}} & 
\multicolumn{1}{>{\columncolor[gray]{0.8}}c}{\textcolor{red}{\textbf{0.1730}}} & 
\multicolumn{1}{>{\columncolor[gray]{0.8}}c}{\textcolor{red}{\textbf{19.35}}} & 
\multicolumn{1}{>{\columncolor[gray]{0.8}}c}{4.7391} & 
\multicolumn{1}{>{\columncolor[gray]{0.8}}c}{0.4538} & 
\multicolumn{1}{>{\columncolor[gray]{0.8}}c}{\textcolor{blue}{\underline{68.21}}} & 
\multicolumn{1}{>{\columncolor[gray]{0.8}}c}{\textcolor{red}{\textbf{0.7007}}}  \\
\midrule
\midrule
\multirow{11}{*}{\textit{RealSR}} 
& BSRGAN  & \textcolor{blue}{\underline{26.38}} & \textcolor{red}{\textbf{0.7655}} & \textcolor{red}{\textbf{0.2656}} & 0.2124 & 141.28 & 5.6356 & 0.3799 & 63.29 & 0.5114 \\
& Real-ESRGAN & 25.65 & 0.7603 & 0.2727 & 0.2065 & 136.33 & 5.8554 & 0.3765 & 60.45 & 0.4518 \\
\cmidrule(lr){2-11}
& LDM-100 & 26.33 & 0.6986 & 0.4148 & 0.2454 & 143.35 & 6.3368 & 0.3841 & 55.82 & 0.5060 \\
& StableSR-200 & 24.60 & 0.7047 & 0.3068 & 0.2163 & 132.20 & 5.7848 & 0.4336 & 65.71 & 0.6298 \\
& PASD-20 & \textcolor{red}{\textbf{26.56}} & \textcolor{blue}{\underline{0.7636}} & 0.2838 & \textcolor{blue}{\underline{0.1999}} & 120.94 & 5.8052 & 0.3887 & 59.89 & 0.4924 \\
& DiffBIR-50  & 24.24 & 0.6650 & 0.3469 & 0.2300 & 134.65 & 5.4909 & 0.4853 & 64.25 & 0.6543 \\
& SeeSR-50 & 25.14 & 0.7210 & 0.3007 & 0.2224 & \textcolor{blue}{\underline{125.44}} & \textcolor{blue}{\underline{5.3971}} & \textcolor{red}{\textbf{0.5429}} & \textcolor{red}{\textbf{69.81}} & 0.6698 \\
\cmidrule(lr){2-11}
& ResShift-15 & \textcolor{blue}{\underline{26.38}} & 0.7567 & 0.3159 & 0.2433 & 149.66 & 6.8703 & 0.3970 & 60.21 & 0.5488 \\
& SinSR-1 & 26.16 & 0.7368 & 0.3075 & 0.2332 & 136.47 & 6.0054 & 0.4035 & 60.95 & 0.6304 \\
& OSEDiff-1 & 24.43 & 0.7153 & 0.3173 & 0.2363 & 126.13 & 6.3821 & \textcolor{blue}{\underline{0.4878}} & 67.53 & \textcolor{red}{\textbf{0.6733}} \\
& \multicolumn{1}{>{\columncolor[gray]{0.8}}l|}{S3Diff-1 (Ours)} &
\multicolumn{1}{>{\columncolor[gray]{0.8}}c}{25.03} &  
\multicolumn{1}{>{\columncolor[gray]{0.8}}c}{0.7321} &  
\multicolumn{1}{>{\columncolor[gray]{0.8}}c}{\textcolor{blue}{\underline{0.2699}}} &  
\multicolumn{1}{>{\columncolor[gray]{0.8}}c}{\textcolor{red}{\textbf{0.1996}}} &  
\multicolumn{1}{>{\columncolor[gray]{0.8}}c}{\textcolor{red}{\textbf{108.88}}} &  
\multicolumn{1}{>{\columncolor[gray]{0.8}}c}{\textcolor{red}{\textbf{5.3311}}} &  
\multicolumn{1}{>{\columncolor[gray]{0.8}}c}{0.4563} &  
\multicolumn{1}{>{\columncolor[gray]{0.8}}c}{\textcolor{blue}{\underline{67.89}}} &  
\multicolumn{1}{>{\columncolor[gray]{0.8}}c}{\textcolor{blue}{\underline{0.6722}}} \\
\midrule
\midrule
\multirow{11}{*}{\textit{DrealSR}} 
& BSRGAN & \textcolor{blue}{\underline{28.70}} & \textcolor{blue}{\underline{0.8028}} & \textcolor{blue}{\underline{0.2858}} & 0.2144 & 155.63 & 6.5296 & 0.3435 & 57.17 & 0.5094 \\
& Real-ESRGAN & 28.61 & \textcolor{red}{\textbf{0.8052}} & \textcolor{red}{\textbf{0.2819}} & \textcolor{red}{\textbf{0.2089}} & 147.62 & 6.6782 & 0.3449 & 54.27 & 0.4520 \\
\cmidrule(lr){2-11}
& LDM-100 & 28.70 & 0.7409 & 0.4849 & 0.2889 & 164.80 & 8.0084 & 0.3240 & 54.35 & 0.6047 \\
& StableSR-200 & 28.24 & 0.7596 & 0.3149 & 0.2234 & 149.18 & 6.6920 & 0.3697 & 57.42 & 0.6062 \\
& PASD-20 & \textcolor{red}{\textbf{29.07}} & 0.7921 & 0.3146 & 0.2179 & \textcolor{blue}{\underline{138.47}} & 7.4215 & 0.3637 & 50.34 & 0.5112 \\
& DiffBIR-50 & 25.93 & 0.6526 & 0.4518 & 0.2762 & 177.41 & \textcolor{blue}{\underline{6.2261}} & \textcolor{blue}{\underline{0.4922}} & 63.47 & 0.6859 \\
& SeeSR-50 & 28.07 & 0.7684 & 0.3174 & 0.2315 & 147.41 & 6.3807 & \textcolor{red}{\textbf{0.5145}} & \textcolor{red}{\textbf{65.08}} & 0.6903 \\
\cmidrule(lr){2-11}
& ResShift-15 & 28.69 & 0.7875 & 0.3525 & 0.2542 & 176.57 & 7.8754 & 0.3505 & 52.37 & 0.5402 \\
& SinSR-1 & 28.37 & 0.7486 & 0.3684 & 0.2475 & 172.77 & 7.0558 & 0.3829 & 54.97 &  0.6333 \\
& OSEDiff-1 & 27.65 & 0.7743 & 0.3177 & 0.2366 & 141.95 & 7.2915 & 0.4845 & 63.55 & \textcolor{blue}{\underline{0.7056}} \\
& \multicolumn{1}{>{\columncolor[gray]{0.8}}l|}{S3Diff-1 (Ours)} & 
\multicolumn{1}{>{\columncolor[gray]{0.8}}c}{26.89} & 
\multicolumn{1}{>{\columncolor[gray]{0.8}}c}{0.7469} & 
\multicolumn{1}{>{\columncolor[gray]{0.8}}c}{0.3122} & 
\multicolumn{1}{>{\columncolor[gray]{0.8}}c}{\textcolor{blue}{\underline{0.2120}}} & 
\multicolumn{1}{>{\columncolor[gray]{0.8}}c}{\textcolor{red}{\textbf{119.86}}} & 
\multicolumn{1}{>{\columncolor[gray]{0.8}}c}{\textcolor{red}{\textbf{6.1647}}} & 
\multicolumn{1}{>{\columncolor[gray]{0.8}}c}{0.4508} & 
\multicolumn{1}{>{\columncolor[gray]{0.8}}c}{\textcolor{blue}{\underline{64.19}}} & 
\multicolumn{1}{>{\columncolor[gray]{0.8}}c}{\textcolor{red}{\textbf{0.7122}}}  \\
\midrule
\midrule
\end{tabular}%
\end{table*}

\section{Experiments} \label{sec:experiments}

\subsection{Experimental Setup} 

\paragraph{Training Details.}
Following the recent work SeeSR \citep{wu2023seesr}, we train our model on the LSDIR~\citep{li2023lsdir} dataset and a subset of 10k face images from FFHQ~\citep{ffhq}. To synthesize the LR-HR pairs for training, we employ the degradation pipeline proposed in Real-ESRGAN \citep{wang2021real}. During training, the synthesized LR images are upscaled to match the HR resolution of $512 \times 512$ before feeding into our SR model. The training process takes over 30k iterations, with a batch size of 64 and a learning rate of $2\text{e}^{-5}$.
\begin{table*}[!t]
\setlength{\tabcolsep}{2.5pt}
\centering
\caption{Quantitative comparison with state-of-the-art methods on RealSet65. The best and second-best results are highlighted in
\textcolor{red}{\textbf{red}} and \textcolor{blue}{\underline{blue}}, respectively.}
\label{tab:compare_methods2}
\vspace{-2mm}
\fontsize{9pt}{12pt}\selectfont
\begin{tabular}{@{}c|cc|ccc|cccc@{}}
\toprule
\toprule
Metrics & BSRGAN & \begin{tabular}[c]{@{}c@{}}Real-\\ ESRGAN\end{tabular} & StableSR-200  & DiffBIR-50 & SeeSR-50 & ResShift-15 & SinSR-1 & OSEDiff-1 & S3Diff-1 \\ \midrule
NIQE $\downarrow$ & 4.5801 & 4.3487 & 5.0371 & \textcolor{blue}{\underline{4.2559}} & 4.8959 & 6.0893 & 5.9221 & 5.2142 & \textcolor{red}{\textbf{4.2467}} \\
MANIQA $\uparrow$ & 0.3898 & 0.3934 & 0.4442 & \textcolor{red}{\textbf{0.4984}} & \textcolor{blue}{\underline{0.4982}} & 0.4106 & 0.4341 & 0.4666 & 0.4685 \\
MUSIQ $\uparrow$ & 65.58 & 64.12 & 62.79 & \textcolor{red}{\textbf{69.81}} & 68.88 & 60.90 & 62.85 & 68.48 & \textcolor{blue}{\underline{68.92}} \\
CLIPIQA $\uparrow$ & 0.6159 & 0.6074 & 0.5982 & \textcolor{red}{\textbf{0.7468}} & 0.6805 & 0.6522 & \textcolor{blue}{\underline{0.7172}} & 0.7061 & 0.7120 \\
\bottomrule
\bottomrule
\end{tabular}
\end{table*}

We adopted SD-turbo~\citep{sauer2023adversarial} as the base T2I model and fine-tuned it using the proposed degradation-guided LoRA. The rank parameter in LoRA is set as $16$ for the VAE encoder and $32$ for the diffusion UNet, respectively.
The hyper-parameters of $\lambda_\text{L2}$, $\lambda_\text{LPIPS}$, and $\lambda_\text{GAN}$ are set to be 2.0, 5.0, 0.5, respectively.
The probability $p_n$ for online negative prompting is set as 0.05.
\begin{table*}[!t]
    \centering
    \caption{Comparison of efficiency across various methods. The inference time of each method is calculated using the average inference time of 3000 input images of size 128 $\times$ 128, upscaled by a factor of 4, on an A100 GPU.}
    \label{tab:efficiency_comparison}
    \vspace{-2mm}
    \begin{tabular}{l|ccccccc}
        \toprule
        \toprule
         & StableSR & DiffBIR & SeeSR & ResShift & SinSR & OSEDiff & \name \\
        \midrule
        Inference Step & 200 & 50 & 50 & 15 & 1 & 1 & 1 \\
        Inference Time (s) & 17.75 & 16.06 & 5.64 & 1.65 & 0.42 & 0.60 & 0.62 \\
        \# Trainable Param (M) & 150.0 & 380.0 & 749.9 & 118.6 & 118.6 & 8.5 & 34.5 \\
        \bottomrule
        \bottomrule
    \end{tabular}
\end{table*}

\paragraph{Testing Details.}
We mainly follow the testing configurations of StableSR \citep{wang2023exploiting} to comprehensively evaluate the performance of the proposed \name. 
We adopt the degradation pipeline of Real-ESRGAN \citep{wang2021real} to create 3,000 LR-HR pairs from the DIV2K validation set \citep{div2k}. The resolution of LR images is 128$\times$128, and the corresponding HR images 512$\times$512. For real-world datasets, RealSR \citep{realsr} and DRealSR, we follow the standard practices of StableSR \citep{wang2023exploiting} and SeeSR \citep{wu2023seesr} by cropping the LR image to 128$\times$128. Besides, we test our method on the real-world dataset, RealSet65 \citep{yue2023resshift}, to comprehensively evaluate the generative ability of our method. The guidance scale $\lambda_{\mathrm{cfg}}$ for Classifier-Free Guidance is set to 1.1.
\begin{figure*}[!t]
    \centering
    \includegraphics[width=2\columnwidth]{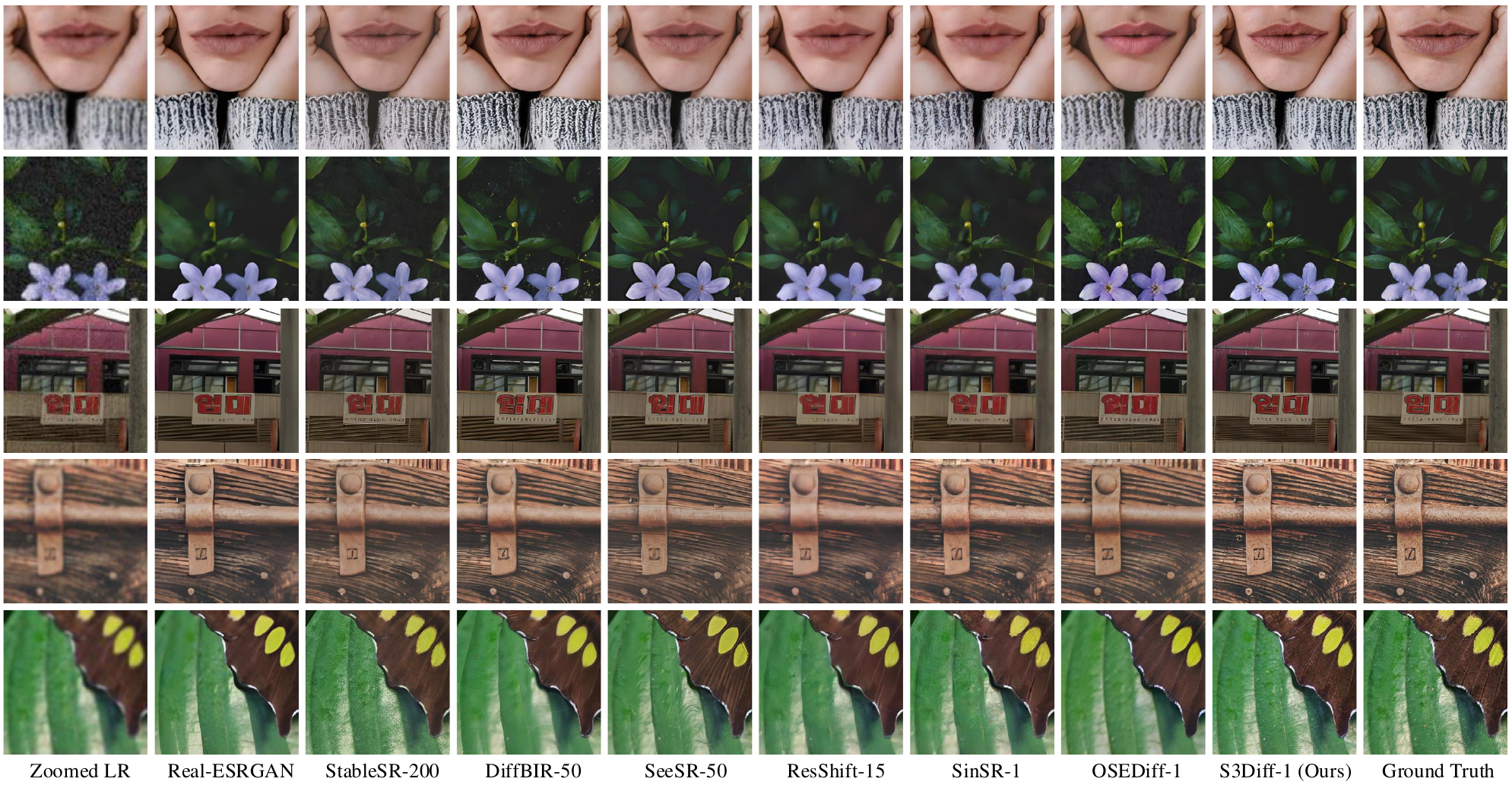}
    \vspace{-2mm}
    \caption{Qualitative comparisons of different methods on the synthesis dataset, \textit{DIV2K-Val \citep{div2k}}. \textbf{(Zoom in for details)}}\label{fig:comparison_syn}
\end{figure*}

\paragraph{Evaluation Metrics.}

To thoroughly evaluate the performance of various methods, we consider both reference and non-reference metrics. PSNR and SSIM \citep{ssim} are reference-based fidelity measures, which are calculated on the Y channel of the YCbCr space. LPIPS \citep{lpips} and DISTS \citep{ding2020image} serve as reference-based perceptual quality measures. FID \citep{heusel2017gans} assesses the distribution distance between ground truth and restored images. Additionally, we use NIQE \citep{zhang2015feature}, MANIQA \citep{yang2022maniqa}, MUSIQ \citep{musiq}, and CLIPIQA \citep{clipiqa} as non-reference metrics to assess image quality.

\paragraph{Compared Methods.}

We compare our method with various cutting-edge real SR methods, which we have grouped into three categories. The first category includes GAN-based methods, including BSRGAN \citep{zhang2021designing} and Real-ESRGAN \citep{wang2021real}. The second category features diffusion-based methods like LDM \citep{rombach2022high}, StableSR \citep{wang2023exploiting}, PASD \citep{yang2023pixel}, DiffBIR \citep{lin2023diffbir}, and SeeSR \citep{wu2023seesr}. The third category comprises state-of-the-art diffusion-based methods with few inference steps, including ResShift \citep{yue2023resshift,yue2024efficient}, SinSR \citep{wang2023sinsr} and OSEDiff \citep{wu2024one}. For a fair comparison, we use their publicly available codes and checkpoints to generate HR images on the same testing sets and report the corresponding performance comparisons.

\subsection{Comparisons with State-of-the-Arts}
\paragraph{Quantitative Comparisons}
We first show the quantitative comparison results on three synthetic and real-world datasets in Table~\ref{tab:compare_methods1}. We can obtain the following observations. 
(1) our approach consistently achieves promising reference metrics. Compared to the accelerated diffusion-based methods, \textit{i.e.}, ResShift, SinSR, and OSEDiff, our S3Diff obtains comparable or better PSNR and SSIM scores. Note that, ResShift and SinSR show better PSNR and SSIM scores. This is primarily because they learn the diffusion process on the residual of LR-HR pairs, requiring training the diffusion model from scratch, rather than utilizing a pre-trained text-to-image model.
(2) the proposed \name excels in perceptual quality, achieving top-tier LPIPS, DISTS, and FID scores across all datasets. For instance, on the DIV2K dataset, we achieve scores of 0.2571 for LPIPS, 0.1730 for DISTS, and 19.35 for FID, which are significantly lower than those of competing methods. This demonstrates our ability to produce visually appealing results that align closely with human perception.
(3) our method consistently outperforms various datasets in non-reference image quality metrics, such as NIQE, MANIQA, MUSIQ, and CLIPIQA. Notably, compared to the state-of-the-art method, SeeSR \citep{wu2023seesr}, our approach achieves superior performance on non-reference metrics while requiring significantly less inference time.
(4) the robust performance across different datasets shows the adaptability and generalization capability of our method, proving its effectiveness in both synthetic and real-world scenarios.
Overall, our method surpasses other diffusion-based methods by achieving significantly better scores on both reference and non-reference metrics, requiring only one-step inference.

Table~\ref{tab:compare_methods2} presents the comparison on RealSet65 \citep{yue2023resshift}. Since RealSet65 has no ground-truth HR images, we only report non-reference metrics. We nearly achieve the best performance among efficient DM-based SR models, notably attaining a NIQE score of 4.2467, which significantly surpasses all other methods.

\paragraph{Complexity Analysis}
Table~\ref{tab:efficiency_comparison} compares the number of parameters of different DM-based SR models and their inference time. The inference time of each method is calculated using an average inference time of 3000 input images of size 128 $\times$ 128, upscaled by a factor of 4, on an A100 GPU.
By utilizing a single-step forward pipeline, \name{} significantly outperforms multi-step methods in inference time. Specifically, \name{} is approximately 30 times faster than StableSR, 9 times faster than SeeSR, and 3 times faster than ResShift. While our method is slightly slower than SinSR and OSEDiff, it achieves superior SR quality. Regarding the number of parameters, our method is only larger than OSEDiff. However, OSEDiff utilizes a large pre-trained degradation-robust tag model with 1407M parameters, originally developed from SeeSR \citep{wu2023seesr}. In contrast, our method employs a lightweight degradation estimation model with only 2.36M parameters. The feature of not requiring image content descriptions makes our method extremely efficient while achieving promising performance.
\begin{figure*}[!t]
    \centering
    \includegraphics[width=1.98\columnwidth]{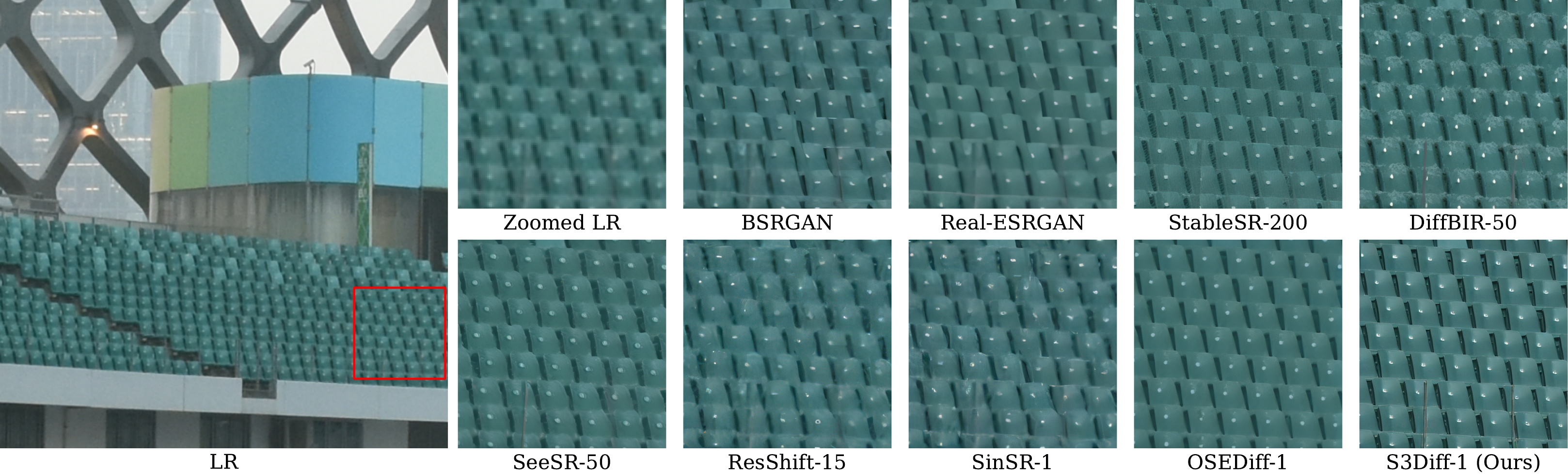}
    \vspace{3mm}
    \includegraphics[width=1.98\columnwidth]{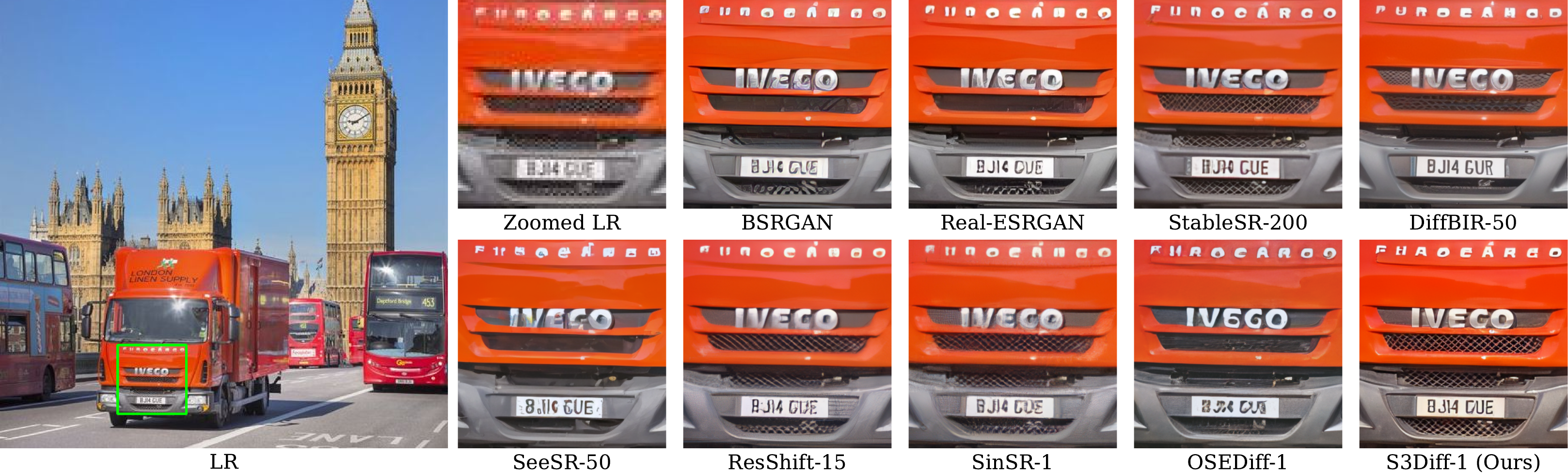}
    \vspace{-3mm}
    \caption{Qualitative comparisons of different methods on the real-world dataset. \textbf{(Zoom in for details)}}\label{fig:comparison_real}
    \vspace{-3mm}
\end{figure*}

\paragraph{Qualitative Comparisons}
Figures~\ref{fig:comparison_syn} and \ref{fig:comparison_real} present visual comparisons of synthetic and real-world images, respectively. 
In Figure~\ref{fig:comparison_syn}, GAN-based methods like BSRGAN and Real-ESRGAN struggle to preserve fine details, resulting in a loss of texture and clarity. DM-based methods, such as StableSR, DiffBIR, and SeeSR, while enhancing detail, often produce outputs inconsistent with the original low-resolution input, leading to unnatural appearances, particularly noticeable in the flowers and leaves. Inference-efficient approaches like ResShift and SinSR, trained from scratch, tend to introduce artifacts that disrupt image smoothness and quality. OSEDiff, in its attempt to enhance details, often causes distortion, resulting in unnatural colors and features.
In contrast, our method excels at accurately reconstructing image features while maintaining semantic integrity, delivering high levels of detail, such as the leaf veins and chapped lips, without introducing artifacts. This underscores our approach's robustness in handling severe degradation and producing reconstructions that are both precise and true to the original content.
\begin{table*}[!t]
\centering
\caption{Quantitative results of different settings for model adaption of our method on \textit{DIV2K-Val}~\citep{div2k} and \textit{RealSR} \citep{realsr} benchmarks.} \label{tab:model_adaption}
\begin{tabular}{l|ccc|ccc}
\toprule
\toprule
\multirow{2}{*}{Module Adaption Methods} & \multicolumn{3}{c|}{\textit{DIV2K-Val}} & \multicolumn{3}{c}{\textit{RealSR}} \\ 
\cmidrule(lr){2-7}
& PSNR $\uparrow$ &  LPIPS$\downarrow$ & MUSIQ $\uparrow$ 
& PSNR $\uparrow$ &  LPIPS$\downarrow$ & MUSIQ $\uparrow$ \\
\midrule
UNet-Lora  & 24.10 & 0.2746 & 63.23 & 24.56 & 0.2770 & 62.42 \\
UNet-Lora + Skip Connection~\citep{parmar2024one} & 24.57 & \textbf{0.2546} & 59.44 & 25.45 & 0.2577 & 57.54 \\
UNet-Lora + Decoder-Lora & 24.25 & 0.2587 & 60.68 & 25.65 & 0.2645 & 54.47 \\
UNet-Lora + Encoder-Lora + Decoder-Lora & \textbf{24.65} & 0.2543 & 53.85 & \textbf{25.82} & \textbf{0.2530} & 54.73 \\ 
\midrule
UNet-Lora + Encoder-Lora (Ours) & 24.12 & 0.2549 & \textbf{65.34} & 24.71 & 0.2751 & \textbf{63.78} \\ 
\bottomrule
\bottomrule
\end{tabular}
\end{table*}

In Figure~\ref{fig:comparison_real}, similar observations can be made for real-world images. GAN-based methods always produce images with distorted structures. Diffusion-based methods like StableSR can produce realistic textures but often struggle to preserve accurate semantic details. Although SeeSR is capable of generating tags to describe image content, sometimes results in inaccurate descriptions, leading to smooth edges and unclear semantic details. While ResShift and SinSR demonstrate better consistency between low-resolution and high-resolution images, they still fall short in capturing fine details. In contrast, our method delivers superior visual results, offering sharp and semantically accurate details, as evidenced by the clear edges and consistent structures throughout the images.

\subsection{Ablation Study} \label{sec:ablation_study}
We first discuss the effectiveness of the proposed strategy of model adaption. Then, we discuss the effectiveness of the proposed degradation-guided LoRA, including its degradation-aware ability and the roles of block ID embeddings. Finally, we investigate the effect of the proposed online negative prompting and ablate on the used losses. Unless stated otherwise, we mainly conduct experiments on the \textit{DIV2K-Val}~\citep{div2k} and \textit{RealSR}~\citep{realsr} datasets.
\begin{table*}[!t]
\centering
\caption{Quantitative results of different settings for degradation-guided LoRA on \textit{DIV2K-Val}~\citep{div2k} and \textit{RealSR} \citep{realsr} benchmarks.} \label{tab:dg_lora}
\begin{tabular}{l|ccc|ccc}
\toprule
\toprule
\multirow{2}{*}{Methods} & \multicolumn{3}{c|}{\textit{DIV2K-Val}} & \multicolumn{3}{c}{\textit{RealSR}} \\ 
& PSNR $\uparrow$ &  LPIPS$\downarrow$ & MUSIQ $\uparrow$ 
& PSNR $\uparrow$ &  LPIPS$\downarrow$ & MUSIQ $\uparrow$ \\
\midrule
Cross-Attention Injection & \textbf{24.67} & 0.2627 & 64.68 & \textbf{25.72} & 0.2675 & 63.67 \\
\midrule
Shared $\bm{C}$  & 24.06 & 0.2613 & 65.52 & 25.66 & 0.2626 & 64.23 \\
Ours (Unshared $\bm{C}$, \emph{w} block ID embeddings) & 24.13 & \textbf{0.2563} & \textbf{66.55} & 25.55 & \textbf{0.2573} & \textbf{65.78} \\ 
\bottomrule
\bottomrule
\end{tabular}
\end{table*}

\paragraph{Effectiveness of Adaption Solution.}
In Table \ref{tab:model_adaption}, we present four experiments to validate the effectiveness of our adaptation solution. We use the same loss functions as in our default setting. To avoid interference from other modules, we do not use degradation-guided LoRA, online negative prompting, and CFG during inference.
\textbf{(a)} We only inject LoRA layers into the UNet. \textbf{(b)} Based on (a), we additionally add several skip connections between the encoder and decoder, which is proposed in~\citep{parmar2024one} to improve input-output structural consistency. \textbf{(c)} We inject LoRA layers into the VAE decoder. \textbf{(d)} Based on (c), we further inject LoRA layers into the VAE encoder. As we can see, injecting LoRAs into the UNet alone can already provide acceptable SR performance, especially perceptual quality. Adding skip connections between the encoder and decoder enhances the information flow, leading to improved structural consistency. However, this setup slightly improves the reference metrics but significantly worsens the non-reference metrics, like the MUSIQ score. This indicates a trade-off between structural similarity and perceptual quality, aligning with findings from StableSR \citep{wang2023exploiting}. Similarly, adding LoRA layers to the VAE decoder results in better structural similarity but reduced perceptual quality, possibly disrupting the well-constructed compressed latent space.
Finally, we try to inject LoRA layers into all modules of the diffusion model, including the VAE encoder, decoder and UNet, leading to the best PSNR score but an unacceptable MUSIQ score. Building on these findings, our method opts to inject LoRA layers only into the VAE encoder and UNet. The encoder LoRA layers help initially recover the LR image, while the LoRA layers of UNet unleash its generative power to polish image details and textures. In the following experiments, we default inject LoRA layers in the VAE encoder and the UNet.
\begin{table*}[!t]
\centering
\caption{Quantitative results of different settings for the proposed online negative prompting on \textit{DIV2K-Val}~\citep{div2k} and \textit{RealSR} \citep{realsr} benchmarks. CFG refers to Classifier-Free Guidance, with guidance scales set to 1.1 for all experiments.} \label{tab:online_negative_prompting}
\begin{tabular}{ll|ccc|ccc}
\toprule
\toprule
\multicolumn{2}{l|}{\multirow{2}{*}{Methods}} & \multicolumn{3}{c|}{\textit{DIV2K-Val}} & \multicolumn{3}{c}{\textit{RealSR}} \\ 
& & PSNR$\uparrow$ &  LPIPS$\downarrow$ & MUSIQ$\uparrow$ 
& PSNR$\uparrow$ &  LPIPS$\downarrow$ & MUSIQ$\uparrow$ \\
\midrule
\multirow{2}{*}{\emph{w/o} Online Negative Prompting} & w/o CFG & \textbf{24.13} & 0.2563 & 66.55 & 25.55 & 0.2573 & 65.78 \\
& \emph{w} CFG & 23.98 & 0.2595 & 67.14 & 25.28 & 0.2610 & 66.25 \\
\midrule
\multirow{2}{*}{\emph{w} Online Negative Prompting} & \emph{w/o} CFG & 24.10 & \textbf{0.2521} & 66.94 & \textbf{25.76} & \textbf{0.2522} & 66.42 \\
& \emph{w} CFG & 23.40 & 0.2571 & \textbf{68.21} & 25.03 & 0.2699 & \textbf{67.88} \\
\bottomrule
\bottomrule
\end{tabular}
\end{table*}

\paragraph{Effectiveness of Degradation-Guided LoRA.} 
In Table \ref{tab:dg_lora}, we conduct three experiments to demonstrate the effectiveness of degradation-guided LoRA. To prevent interference, we avoid using online negative prompting.
In our method, we introduce the auxiliary information of image degradation into the model by modulating the LoRA's weights. There are some other ways of incorporating image degradation, including the use of cross-attention \citep{chen2023image,gandikota2024text} in the way of textual prompt injection.
Therefore, we first explore this method of injecting image degradation. We initialize two degradation prompts using the text embeddings of ``noise'' and ``blur'' from the CLIP text encoder. These prompts are then incorporated into the template: \emph{``This image contains} \textbf{[n]} \emph{and} \textbf{[b]} \emph{degradation''}, where \textbf{[n]} and \textbf{[b]} serve as placeholders for the degradation prompt, which are modulated by the estimated noise and blur scores. As shown in Table \ref{tab:dg_lora}, our solution outperforms the cross-attention-based solution, indicating the effectiveness of incorporating degradation into LoRAs.
We then investigate different implementations of the degradation-guided LoRA. 
Compared with the vanilla LoRA (ours in Table \ref{tab:model_adaption}), we find improvements with the degradation-guided LoRA, even when we use the shared modulation matrix $\bm{C}$. Incorporating the block ID embeddings can improve the flexibility of the model to handle the complexity of image degradation, further enhancing the performance.

\paragraph{Effectiveness of Online Negative Prompting.}

\begin{figure}[!t]
    \centering
    \includegraphics[width=1.\columnwidth]{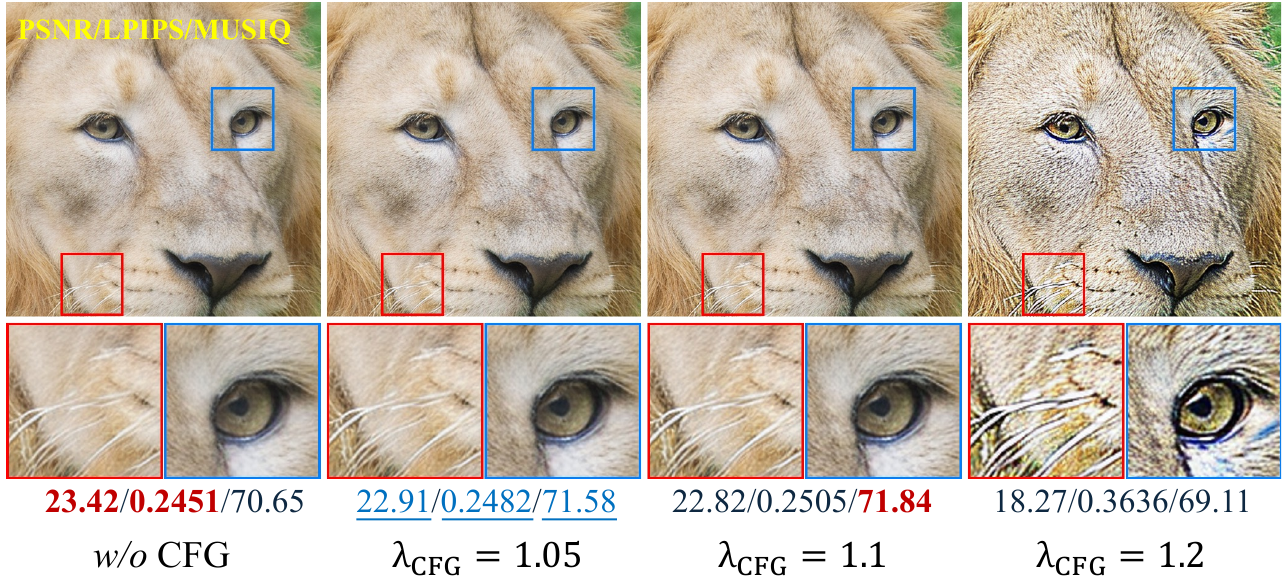}
    \caption{Qualitative comparisons of different guidance scales using CFG. (\textbf{Zoom in for details})}\label{fig:cfg_coef}
\end{figure}

To validate the effectiveness of the proposed online negative prompting, we present four experiments in Table~\ref{tab:online_negative_prompting}, where we also employ the degradation-guided LoRA.
Notably, we discovered that \name can respond to negative prompts even without training, likely due to the inherent capability of the diffusion prior. As shown in Table~\ref{tab:online_negative_prompting}, when trained only with positive samples and tested using CFG, the perceptual quality of generated images slightly improves compared to not using CFG. After incorporating negative samples, both structural similarity and perceptual quality are improved, indicating the SR model recognizes what constitutes good and poor image quality through the proposed online negative prompting. In this context, using CFG during inference significantly boosts non-reference metric scores, making the slight reduction in reference metric scores worthwhile. As shown in Figure~\ref{fig:cfg_coef}, the guidance scale greatly affects the final outcome, which may be because SD-Turbo~\citep{sauer2023adversarial} does not support CFG during testing. Therefore, we set it to a small value of 1.1 to balance fidelity and perceptual quality.

\paragraph{Impact of LoRA Rank.}

LoRA plays a crucial role in our approach, with the LoRA rank being a key hyper-parameter. In Table \ref{tab:lora_rank}, we assess the impact of various LoRA ranks on super-resolution performance. As we can see, a low rank primarily affects reference metrics like PSNR and LPIPS. Using a low LoRA rank, such as 4 or 8, results in unstable training and poor structural similarity. This may be due to the domain gap between SR and T2I, which requires enough trainable parameters to bridge effectively. In contrast, non-reference metrics, like MUSIQ, are only slightly affected by low ranks, possibly due to the intrinsic generative ability of the T2I diffusion model. However, a higher rank, such as 32, might reach a saturation point where results no longer improve. We find that a rank of 32 for the UNet and 16 for the VAE encoder strikes a good balance between model complexity and super-resolution performance.
\begin{table}[!t]
\setlength{\tabcolsep}{1.2pt}
\centering
\caption{Quantitative results of different LoRA ranks on \textit{DIV2K-Val}~\citep{div2k} and \textit{RealSR}~\citep{realsr} benchmarks. 8/16 indicates a setting of 8 for the VAE encoder and 16 for the UNet.} \label{tab:lora_rank}
\fontsize{9pt}{12pt}\selectfont
\begin{tabular}{l|ccc|ccc}
\toprule
\toprule
\multirow{2}{*}{LoRA} & \multicolumn{3}{c|}{\textit{DIV2K-Val}} & \multicolumn{3}{c}{\textit{RealSR}} \\ 
\cmidrule(lr){2-7}
Rank & PSNR$\uparrow$ & LPIPS$\downarrow$ & MUSIQ$\uparrow$ 
& PSNR$\uparrow$ &  LPIPS$\downarrow$ & MUSIQ$\uparrow$ \\
\midrule
4/4 & 22.55 & 0.2720 & 68.28 & 23.99 & 0.2742 & 68.10 \\
8/8 & 22.61 & 0.2723 & 68.41 & 24.16 & 0.2663 & 68.28 \\
8/16 & 23.08 & 0.2690 & 68.11 & 24.22 & 0.2602 & 67.97 \\
16/16 & 23.10 & 0.2640 & 67.69 & 24.31 & 0.2705 & 68.11 \\
16/32 & 23.40 & \textbf{0.2571} & 68.21 & \textbf{25.03} & 0.2699 & 67.88 \\
32/32 & \textbf{23.45} & 0.2577 & \textbf{68.35} & 24.98 & \textbf{0.2687} & \textbf{67.90} \\
\bottomrule
\bottomrule
\end{tabular}
\end{table}
\begin{figure}[!t]
    \centering
    \includegraphics[width=1.\columnwidth]{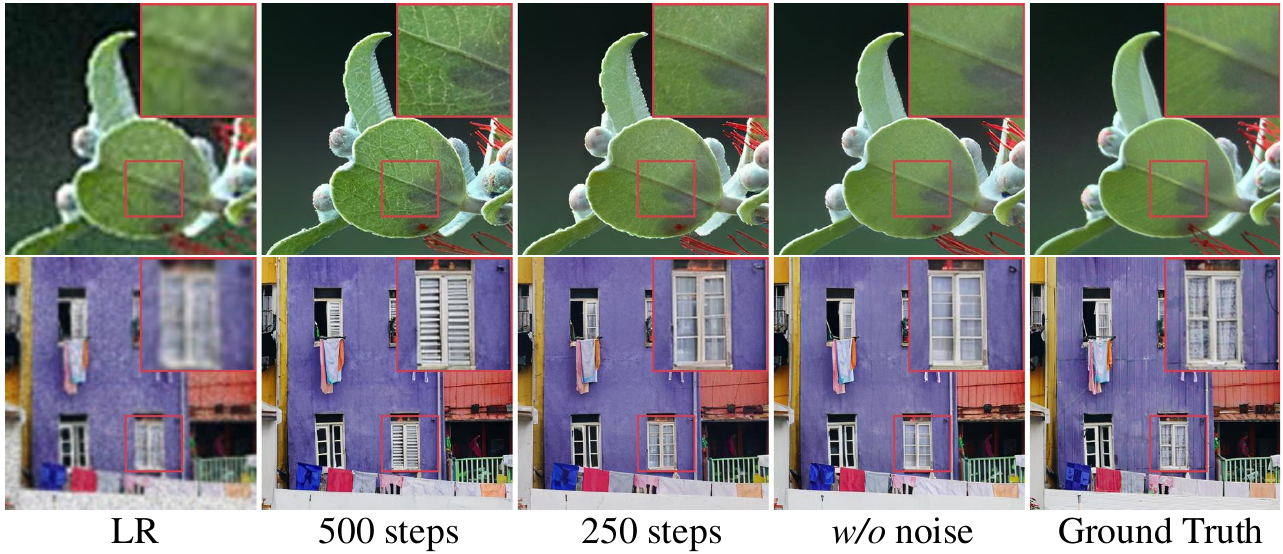}
    \caption{Qualitative comparisons of injecting noise of different levels into LR images as input. (\textbf{Zoom in for details})}\label{fig:start_steps}
\end{figure}
\begin{figure*}[!t]
    \centering
    \includegraphics[width=2.\columnwidth]{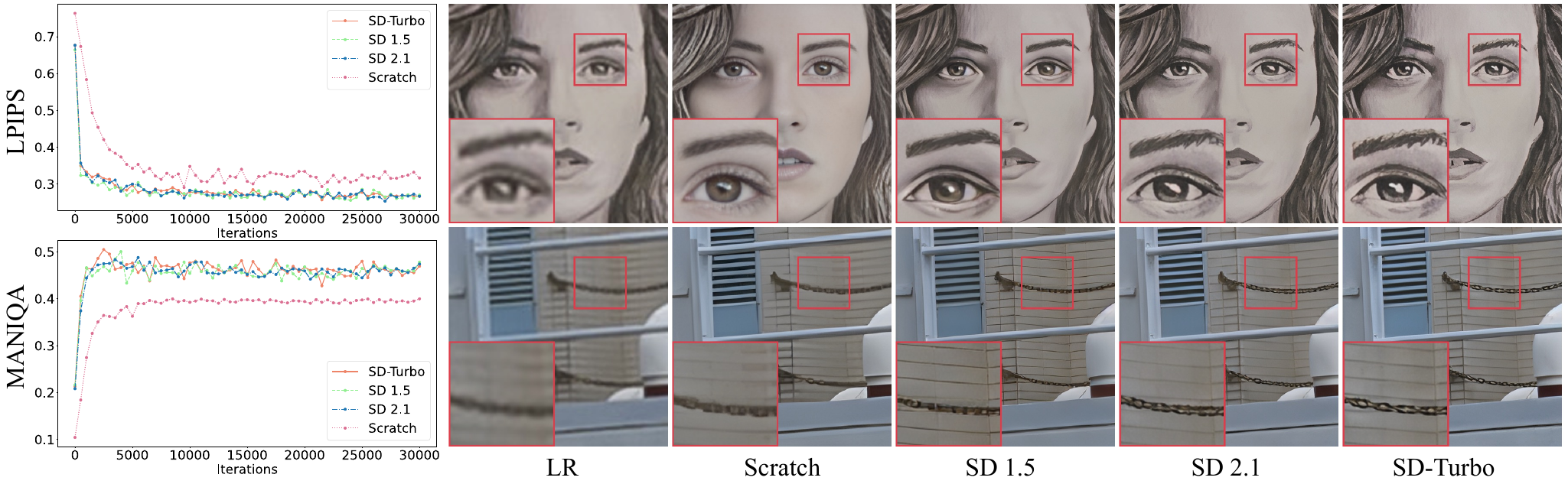}
    \caption{Comparisons of SR performance (LPIPS and MANIQA scores) and convergence speed between different diffusion priors. Visualization results on the RealSR dataset demonstrate the advantages of using diffusion priors. (\textbf{Zoom in for details})}\label{fig:diffusion_prior}
    \vspace{-2mm}
\end{figure*}

\paragraph{Impact of Starting Step.}
From Figure~\ref{fig:start_steps_comparison}, we can find that SD-Turbo can recover the LR image at different noise levels. 
Here, we investigate the impact of the starting step on the SR results. Specifically, we add noise to the LR image at different levels based on the starting step. As shown in Table~\ref{tab:start_steps}, directly using LR images or adding noise at 250 steps yields comparable results. This is because SD-Turbo mainly adds details in the last 250 steps, which does not significantly affect the image structure. In contrast, starting at 500 steps decreases the PSNR score but improves the MUSIQ score. This is reasonable because the image structure of the LR image is significantly degraded at this point. Additionally, the generation capability at 500 steps is much stronger, allowing for the creation of more diverse images. Visual examples in Figure~\ref{fig:start_steps} further demonstrate this finding. To maintain a trade-off between non-reference and reference metrics, we directly use the LR image as input for simplicity.
\begin{table}[!t]
\setlength{\tabcolsep}{1.2pt}
\centering
\caption{Quantitative results of injecting noise of different levels into LR images on \textit{DIV2K-Val}~\citep{div2k} and \textit{RealSR} \citep{realsr} benchmarks.} \label{tab:start_steps}
\fontsize{8pt}{12pt}\selectfont
\begin{tabular}{l|ccc|ccc}
\toprule
\toprule
\multirow{2}{*}{Start} & \multicolumn{3}{c|}{\textit{DIV2K-Val}} & \multicolumn{3}{c}{\textit{RealSR}} \\ 
\cmidrule(lr){2-7}
Step & PSNR$\uparrow$ & LPIPS$\downarrow$ & MUSIQ$\uparrow$ 
& PSNR$\uparrow$ &  LPIPS$\downarrow$ & MUSIQ$\uparrow$ \\
\midrule
500 & 22.50 & 0.2662 & \textbf{68.90} & 23.39 & 0.2877 & \textbf{67.95} \\
250 & 23.12 & 0.2597 & 68.05 & 24.70 & \textbf{0.2673} & 67.52 \\
\emph{w/o} noise & \textbf{23.40} & \textbf{0.2571} & 68.21 & \textbf{25.03} & 0.2699 & 67.88 \\
\bottomrule
\bottomrule
\end{tabular}
\end{table}
\begin{figure}[!t]
  \centering
  \includegraphics[width=0.99\columnwidth]{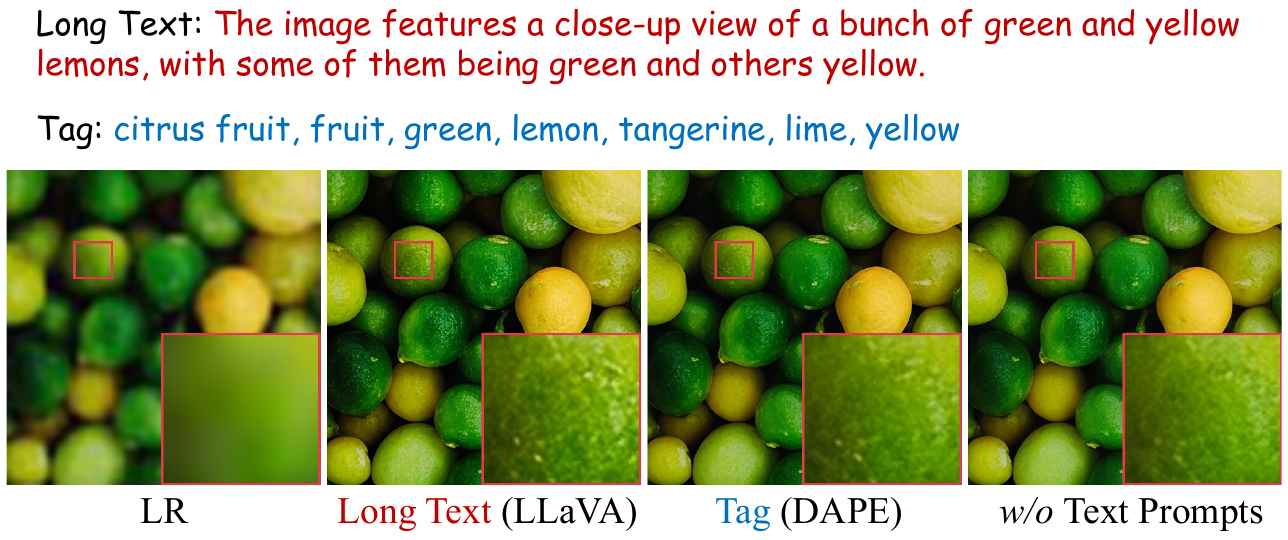}
  \caption{Qualitative comparisons of using different text prompts. (\textbf{Zoom in for details})}\label{fig:plot_text_prompts}
\end{figure}

\paragraph{Impact of Diffusion Prior.}

The default \name{} utilizes SD-Turbo~\citep{sauer2023adversarial}, capable of generating images in just a few sampling steps, to quickly adapt the T2I prior to super-resolution. We first evaluate the effectiveness of using a proper diffusion prior by training a baseline from scratch without loading a pre-trained model.
Notably, since the UNet comprises the majority of trainable parameters, we only train the UNet from scratch and freeze the VAE encoder to conserve GPU memory.
Additionally, we use Stable Diffusion 1.5 and 2.1 as diffusion priors for comparison. The architecture is kept consistent with \name{} to ensure fairness. As illustrated in Figure~\ref{fig:diffusion_prior}, utilizing diffusion priors significantly improves both convergence speed and SR performance compared to training from scratch. Moreover, we observe that training from scratch requires 2$\sim$3 times more GPU memory on average compared to \name. Leveraging the distilled diffusion prior, SD-Turbo slightly enhances convergence speed. Moreover, as demonstrated in Figure~\ref{fig:diffusion_prior}, all three SD models achieve similar SR performance after convergence.
\begin{table}[!t]
\setlength{\tabcolsep}{1.1pt}
\centering
\caption{Quantitative results of integrating different text prompts on \textit{DIV2K-Val}~\citep{div2k} and \textit{RealSR} \citep{realsr} benchmarks.} \label{tab:text_prompt}
\fontsize{8pt}{12pt}\selectfont
\begin{tabular}{l|ccc|ccc}
\toprule
\toprule
\multirow{2}{*}{Method} & \multicolumn{3}{c|}{\textit{DIV2K-Val}} & \multicolumn{3}{c}{\textit{RealSR}} \\ 
\cmidrule(lr){2-7}
& PSNR$\uparrow$ & LPIPS$\downarrow$ & MUSIQ$\uparrow$ 
& PSNR$\uparrow$ &  LPIPS$\downarrow$ & MUSIQ$\uparrow$ \\
\midrule
Ours & \textbf{23.40} & \textbf{0.2571} & 68.21 & \textbf{25.03} & \textbf{0.2699} & 67.88 \\
+ Tag & 23.15 & 0.2653 & 68.41 & 24.86 & 0.2745 & 68.02 \\
+ Long text & 23.28 & 0.2610 & \textbf{69.57} & 24.88 & 0.2772 & \textbf{68.57} \\
\bottomrule
\bottomrule
\end{tabular}
\end{table}

\paragraph{Integrating with Text Prompts.}

Even though the default \name does not utilize text prompts like previous methods~\citep{lin2023diffbir,wu2023seesr,yu2024scaling,wu2024one}, which help enhance image detail recovery, \name can still be seamlessly combined with textual descriptions, such as tags~\citep{wu2023seesr,wu2024one} or long texts~\citep{lin2023diffbir,yu2024scaling}, by substituting the used general positive prompt. Table~\ref{tab:text_prompt} shows the experimental results on three options. 
The first experiment is based on our default setting, \emph{i.e.}, we utilize a general prompt: \emph{``a high-resolution image full of vivid details, showcasing a rich blend of colors and clear textures''}. The second experiment employs the DAPE, which is proposed by SeeSR \citep{wu2023seesr} to extract degradation-robust tag-style prompts. The third experiment uses long text descriptions extracted by LLaVA-v1.5~\citep{liu2024visual}, following SUPIR~\citep{yu2024scaling}. As we can see, using text prompts can improve non-reference metrics but may decrease reference metrics. By employing DAPE and LLaVA to extract text prompts, the generation capability of the pre-trained T2I model is activated, leading to richer synthesized details, though this can reduce structural consistency. A visual example is shown in Figure~\ref{fig:plot_text_prompts}. Although DAPE and LLaVA provide semantic information from the low-resolution image, their visual details are similar to ours. Additionally, they introduce large models with a huge number of parameters for extracting image descriptions. For example, DAPE and LLaVA have 1.4B and 7B parameters, which demand significant hardware resources for inference. More seriously, LLaVA takes 4 to 5 seconds to generate a text prompt per image, complicating efficient real-time super-resolution. Given the comparable performance and to ensure efficient inference, we do not use text prompts in our default setting.

\subsection{Degradation Control}

\begin{figure}[!t]
  \centering
  \includegraphics[width=0.99\columnwidth]{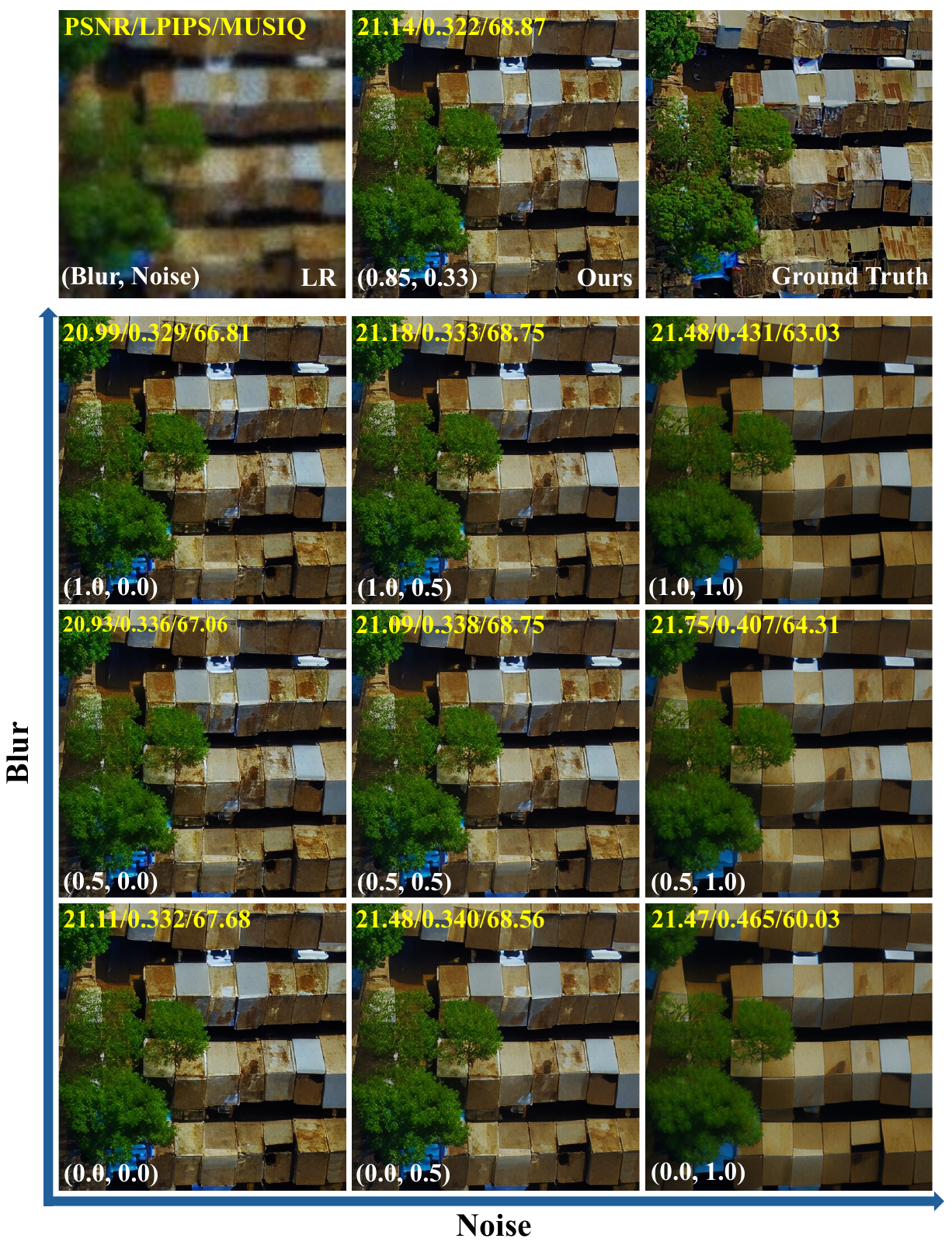}
  \caption{Visual comparison on different degradation inputs. We replace the noise score or the blur score of estimated degradation with 0, 0.5, or 1.0 in our experiments. (\textbf{Zoom in for details})}\label{fig:degradation_control}
\end{figure}

We observe that degradation information significantly impacts SR performance. Here, we aim to figure out whether the model utilizes degradation information. To do this, we manually adjust the predicted noise and blur scores to predefined values and use our model to generate corresponding super-resolution images. Figure \ref{fig:degradation_control} illustrates an example where the blur and noise scores are estimated at 0.85 and 0.33, respectively. As shown, when the input noise score increases, the generated images become smoother. This may reduce perceptual quality but can sometimes enhance the consistency between LR and HR images by eliminating incorrect predictions. Moreover, adjusting the blur score primarily affects the richness of image details. These experiments show our method's capability to utilize degradation information to tailor the output according to users' demands. Compared to images generated with preset scores, the image generated using estimated scores achieves nearly the best metric results, highlighting the effectiveness of our approach in utilizing degradation information.


\section{Limitations}

While \name{} gains advantages from the diffusion prior, it also faces some limitations. Specifically, \name{} struggles with handling small scene texts, human faces, etc. Even if these challenges are common in existing super-resolution methods, we believe that utilizing a more advanced diffusion prior, like SDXL~\citep{podell2023sdxl} and SD3~\citep{esser2024scaling}, and training on higher-quality data could offer improvements. We plan to explore these in future work.

\section{Conclusion}
This work introduced a novel one-step SR model, fine-tuned from a pre-trained T2I model, effectively addressing the efficiency limitations of diffusion-based SR methods. We developed a degradation-guided LoRA module that enhances SR performance by integrating estimated degradation information from LR images while preserving the robust generative priors of the pre-trained diffusion model. This dual focus on efficiency and degradation modeling results in a powerful, data-dependent SR model that significantly outperforms recent state-of-the-art methods. Our approach also includes an online negative prompting strategy during the training phase and classifier-free guidance during inference, greatly improving perceptual quality. Extensive experimental results demonstrate the superior efficiency and effectiveness of our method, achieving high-quality image super-resolution with one sampling step.

\section*{Data Availability Statement}

All experiments are conducted on publicly available datasets. Refer to the references cited.

%
%

\bibliographystyle{spbasic}      
\bibliography{sn-bibliography}   

\end{document}